\pdfoutput=1

\documentclass[11pt]{article}
\usepackage[table,x11names,dvipsnames,table]{xcolor}

\usepackage[final]{acl}

\usepackage{times}
\usepackage{latexsym}

\usepackage{makecell}
\usepackage{adjustbox}

\usepackage{anyfontsize}

\definecolor{anger}{RGB}{141,140,197}
\definecolor{sadness}{RGB}{208,124,139}
\definecolor{fear}{RGB}{23,163,186}
\definecolor{surprise}{RGB}{236,181,91}
\definecolor{optimism}{RGB}{239,151,131}
\definecolor{joy}{RGB}{226,136,110}

\usepackage[T1]{fontenc}

\usepackage[utf8]{inputenc}

\usepackage{microtype}

\usepackage{inconsolata}

\usepackage{graphicx}

\newcommand{\emopiintitle}[4]{
  \begingroup
  \renewcommand{\arraystretch}{0.2}
  \begin{tabular}[b]{@{}c@{}}
    \textcolor{black}{\fontsize{10}{18}\selectfont #1}\textcolor{black}{\fontsize{10}{18}\selectfont #2}\textcolor{black}{\fontsize{10}{18}\selectfont #3}\\
    \textcolor{black}{\fontsize{21}{18}\selectfont #4}
  \end{tabular}
  \endgroup
}

\newcommand{\emopiintext}[2]{
  \begingroup
  \renewcommand{\arraystretch}{0.2}
  \begin{tabular}[b]{@{}c@{}}
    \textcolor{black}{\fontsize{7}{18}\selectfont #1} \\
    \textcolor{black}{\fontsize{15}{18}\selectfont #2}
  \end{tabular}
  \endgroup
}

\newcommand{\emopitable}[2]{
  \begingroup
  \renewcommand{\arraystretch}{0.2}
  \begin{tabular}[b]{@{}c@{}}
    \textcolor{black}{\fontsize{8}{18}\selectfont #1} \\
    \textcolor{black}{\fontsize{15}{18}\selectfont #2}
  \end{tabular}
  \endgroup
}

\newcommand{\meanstd}[2]{
  \begingroup
  \renewcommand{\arraystretch}{0.2}
  \begin{tabular}{@{}c@{}}
    \textcolor{black}{\selectfont #1} \\
    \textcolor{black}{\fontsize{8}{18}\selectfont #2}
  \end{tabular}
  \endgroup
}

\newcommand\Tstrut{\rule{0pt}{2.5ex}}       
\newcommand\Bstrut{\rule[-1.0ex]{0pt}{0pt}} 

\title{Emo Pillars \emopiintitle{e}{m}{o}{$\pi$}: Knowledge Distillation to Support Fine-Grained Context-Aware and Context-Less Emotion Classification}

\author{Alexander Shvets \\
  NLP Group, Pompeu Fabra University, Barcelona, Spain \\
  Language Technologies Unit, Barcelona Supercomputing Center, Spain \\
  \texttt{alexander.shvets@upf.edu, aleksandr.shvets@bsc.es} \\}

\begin{document}
\maketitle
\begin{abstract}
    Most datasets for sentiment analysis lack context in which an opinion was expressed, often crucial for emotion understanding, and are mainly limited by a few emotion categories. Foundation large language models (LLMs) like GPT-4 suffer from over-predicting emotions and are too resource-intensive. We design an LLM-based data synthesis pipeline and leverage a large model, Mistral-7b, for the generation of training examples for more accessible, lightweight BERT-type encoder models. We focus on enlarging the semantic diversity of examples and propose grounding the generation into a corpus of narratives to produce non-repetitive story-character-centered utterances with unique contexts over 28 emotion classes. By running 700K inferences in 450 GPU hours, we contribute with the dataset of 100K contextual and also 300K context-less examples to cover both scenarios. We use it for fine-tuning pre-trained encoders, which results in several Emo Pillars \emopiintext{emo}{$\pi$} models. We show that \emopiintext{emo}{$\pi$} models are highly adaptive to new domains when tuned to specific tasks such as GoEmotions, ISEAR, IEMOCAP, and EmoContext, reaching the SOTA performance on the first three. We also validate our dataset, conducting statistical analysis and human evaluation, and confirm the success of our measures in utterance diversification (although less for the \textit{neutral} class) and context personalization, while pointing out the need for improved handling of out-of-taxonomy labels within the pipeline.
\end{abstract}

\section{Introduction}
Manual tagging of texts with emotion categories is a very burdensome process. It is overwhelmed with a multitude of patternless ways to express the same emotion. At the same time, similar utterances may refer to a different set of emotions, as they are often context- and personality-dependent, which hinders establishing a basis for consistent annotation, especially within a fine-grained manifold \citep{devillers2005challenges}. Most datasets in sentiment analysis contain overgeneralized coarse-grained schemes with rare exceptions in context-less settings that would consider a fair number of labels \citep{tu2022context, lykousas2019sharing, demszky2020goemotions}. However, without context, the task is highly subjective, and the interpretation of a situation depends solely on the annotators' priors, leading to large disagreements \citep{park2021dimensional}.

\begin{figure}[t]
  \centering
  \includegraphics[width=0.85\columnwidth]{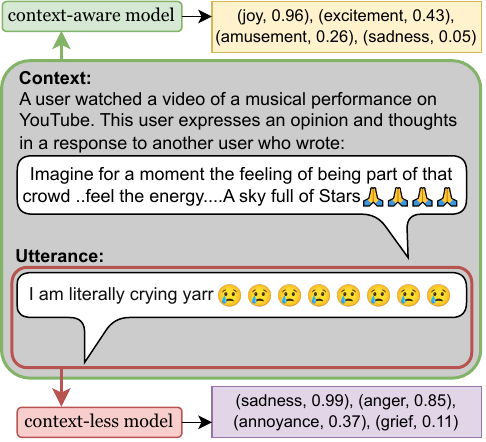}
  \caption{Difference in context-less (context cannot be taken into account) and context-aware (context helps) emotion classification. Context-less models detect emotions in the entire input (including context if provided), while context-aware models can grasp the input structure and extract emotions only from the utterance.}
  \label{fig:example}
\end{figure}

Figure~\ref{fig:example} depicts the importance of context in automatic emotion understanding. Models trained in a regular context-less setup would either ignore the provided situation or extract emotions from the entire input, not being able to distinguish its parts. Foundation large language models that solve various contextual tasks without task-specific tuning (LLMs), e.g. GPT-\{3.5,4\} \citep{openai2024gpt}, can grasp the compound structure of the input, but they understand emotions poorly \citep{wang2024large, chochlakis2024strong, sabour-etal-2024-emobench}.

Synthetic data generation using LLMs has become a common practice in addressing the data scarcity problem. However, it possesses many limitations, with the lack of diversity of the generated data being one of the main issues, especially for high-subjectivity tasks \citep{li-etal-2023-synthetic}. Models tend to generalize too much and produce rather similar items. This requires looking for techniques to make outputs more diverse, such as special ``diversity prompts'', incorporation of real-world data examples within a few-shot setting, and varying the temperature parameter, which can lead to less realistic semantics \citep{li-etal-2023-synthetic, kok2023intertwining, jin2024gpt}. The problem is even more acute for massive inference \citep{yang2023context}.

We propose to use a combination of measures to enhance the semantic diversity of LLMs' outcomes when extracting knowledge to create a large emotion classification dataset: (i) grounding examples into a variety of story texts from topic-rich corpora rather than using generic instructions, (ii) evoking the model to consider each text many times from the perspective of different characters and guess their thoughts and emotions that varied throughout the narrative, and (iii) requesting the model to generate several thoughts/utterances at once while reducing the chances of word repetition at the level of model parameters. In addition, to mitigate the issue of subjectivity, we propose to prompt the model to choose multiple additional ``soft'' co-occurring labels for each example rather than stick to a single ``hard'' one and create utterance-centered summaries of ground texts that would decrease the number of interpretations. Finally, to enhance the robustness of classifiers, we propose to make emotions less explicit in an utterance so that reliance on the context would become a key to understanding them.

To unite these measures, we design an LLM-based data synthesis pipeline that covers many emotions and topics, relates well contexts to utterances, and maintains realistic semantics. We produce and leverage the data to fine-tune mid-sized encoder-based models to \textit{support} context-aware and context-less scenarios encountered in real-world applications. The resulting \textit{Emo Pillars} \emopiintext{emo}{$\pi$} models that are not prone to hallucinations (as they are non-autoregressive) and require less computing than LLMs at inference, reach SOTA performance on several tasks through transfer learning. We release our code\footnote{\href{https://github.com/alex-shvets/emopillars}{https://github.com/alex-shvets/EmoPillars}}, dataset\footnote{\href{https://huggingface.co/datasets/alex-shvets/EmoPillars}{huggingface.co/datasets/alex-shvets/EmoPillars}}, and models\footnote{\href{https://huggingface.co/collections/alex-shvets/emopillars-67ec9694541e0bc69d62861f}{huggingface.co/collections/alex-shvets/EmoPillars}} openly.

\section{Related Work}
Despite the very advanced abilities of LLMs they still can hardly grasp sentiments. \citet{wang2024large} provided a good error analysis showing three main issues when LLMs are applied as-is: hallucination, over-labelling, and over-interpretation. \citet{chochlakis2024strong} added that few-shot learning does not improve the outcome a lot as LLMs are unable to fully integrate information from subjective demonstrations that contrast their strong task priors. Instead, they tend to ossify their predictions; the larger the model, the stronger the effect is observed.

Medium-sized language models, like RoBERTa \citep{liu2019robertarobustlyoptimizedbert}, remain a better option \citep{alvarez2021uncovering, park2021dimensional, zanwar2022improving, cortiz2022exploring, kok2023intertwining}, although they require more training examples per emotion than are usually available in existing datasets. Data augmentation using LLMs has become a common practice in enlarging minor classes. However, it is limited by creating derivatives of the dataset examples through paraphrasing and inspirational generation based on them \citep{wang2024large}, which does not bring much content diversity and may not always align with the expected sentiment \citep{wozniak2023big}. Another type of data enhancement for a given task, e.g. as in \citet{koufakou2023data}, consists in integrating an external dataset such as GoEmotions \citep{demszky2020goemotions}, which is so far the largest third-person annotated fine-grained sentiment dataset of 58K Reddit comments over 28 labels. As it is highly unbalanced, it is either taken only partially with dominant labels \citep{park2021dimensional, zanwar2022improving} or with pre-augmentation, as e.g. in \citet{kok2023intertwining} using GPT-\{3.5,4\} with a simple 5-shot prompt asking to generate 20 records for a provided emotion. Moreover, comments of GoEmotions can be perceived ambiguously in the absence of context, and therefore, alternative labels can be acceptable \citep{kocon2023chatgpt}. The latter happens even with first-party annotated comments \citep{malko2021demonstrating}. To reduce the number of possible interpretations in GoEmotions, \citet{yang2023context} proposed to extend the comments by prompting GPT-\{3.5,4\} to add 1-2 sentences at the end so that they clearly convey the ground truth labels. However, their analysis showed reduced vocabulary diversity as the same 3 words are reused in 20-30\% cases within an emotion. This brings a risk of over-tuning the models to limited patterns.

\section{Data Synthesis Pipeline}
\begin{figure}[t]
  \includegraphics[width=\columnwidth]{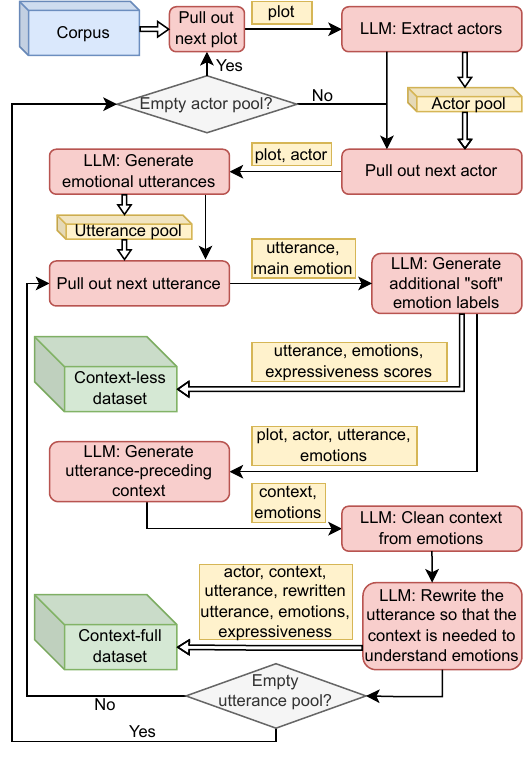}
  \caption{Our pipeline for the generation of a dataset for a multi-label context-aware (-less) emotion recognition.}
  \label{fig:pipeline}
\end{figure}

Figure~\ref{fig:pipeline} depicts our procedure of example generation. It is described in the subsequent subsections, while the implementation details are in Section~\ref{sec:dataset_generation}. We perform prompting in several successive steps to let the model focus on a single task at a time, thus having fewer constraints on generating the next tokens. Since the structure of answers has minor variations for a fixed prompt, we can parse them with simple regex-based extractors. Table~\ref{prompts-table} in Appendix~\ref{sec:appendix_details} lists the designed prompts and corresponding outcomes.

\subsection{Content-rich instructions}
The uniqueness of instructions is reached by iterating over texts in a corpus of narratives (e.g., story plots or news items) and different characters (actors). The actors are identified using an LLM with a simple prompt, as we are not aiming for precise extraction. We iterate over all the found actors to increase the coverage of many diverse emotions and alternative stances raised in the text.

\subsection{Multiple example generation}
We request the model to produce several utterances of the same actor within a single inference and also cover various emotional classes: this gives a model a higher chance of avoiding similar utterances for emotions that can co-occur (i.e., be expressed within the same utterance) and resulting in more contrasting examples where the primary emotion would be presented more expressively. We provide definitions of emotions in the prompt to reduce ambiguity for related labels such as, for example, \textit{sadness} and \textit{disappointment}. Along with emotional utterances, we request two \textit{neutral} ones at the same inference to only slightly anticipate the real-world prevalence of neutral sentiment.

\subsection{Soft labelling}
At this step, we assign multiple emotion labels for an utterance. We eliminate the plot from the prompt so that an LLM does not attend to the actor's emotions in other moments of the story and thus has less information to hallucinate. Instead, we provide the primary emotion for which the utterance was generated. This allows for mitigating the problem of ambiguous sentiment for short utterances. We request the labels to be soft, i.e., with an assessed expressiveness level from 0 to 1 with a step of 0.1. These scores are independent and thus do not sum up to 1. We select labels scored above an established threshold of 0.3 for further steps.

\subsection{Context generation and cleaning}
This step is to reconstruct the situation in which the character arrives to pronounce the utterance. The challenging task is to formulate a prompt so that the LLM does not summarize the entire plot but centers the context around the character and ends it before the moment the utterance is expressed, not revealing the emotions. We found out that the models are sensitive to slight changes in wording in this case. For example, before adding ``Be as concise as possible.'', the output was twice as long and often included information describing post events and emotions\footnote{For example, some context would include a sentence ``He begins to see her as more than just a replicant and develops feelings of caring and protection towards her.''}. To ensure the absence of explicit emotions in contexts, we complemented the pipeline with a context-cleaning step that removes emotive clauses and adjusts affected sentences.

\subsection{Context importance upscale}
\label{subsec:context_importance}
Our context generation aims to provide more clarity to utterances. However, in case an utterance is self-explanatory, i.e., it contains enough signals to interpret emotions, a model may not learn to attend to the information in contexts during the training phase. Then, at prediction time, the attention weights will be kept low for the context tokens even when they are necessary for deriving correct labels. To ensure the importance of the contexts in the dataset, while observing that our generated utterances are rather detailed, we added a step of their rewriting. We specify in the prompt that the context shall become crucial for the emotion understanding by reducing the explicitness of emotions in the utterance. In the following example, the updated utterance drops the markers ``scared'' and ``safely'' that were emphasizing the emotions of \textit{fear} and \textit{caring} correspondingly:

\begin{quote}
\begin{description}
\item \textit{Original}: ``I'm really \underline{scared} right now. I don't know what to do. I need to get this plane down \underline {safely}.''

\item \textit{Rewritten}: ``I'm not sure what to do. I need to land this plane.''
\end{description}
\end{quote}

\noindent If the original utterance expresses emotions mildly, the change is minor and the meaning is kept intact.

\section{Dataset Generation}
\label{sec:dataset_generation}
The seed corpus we use is compiled of 113K synopses of movies, books, and TV shows from English Wikipedia\footnote{\href{https://github.com/markriedl/WikiPlots}{https://github.com/markriedl/WikiPlots}}. According to \citet{papalampidi2022towards}, the stories of this corpus contain intervening events and non-linearities, as well as many characters with elaborate and diverse attributes.

We limit the emotion taxonomy to the 28 GoEmotions categories using definitions from its paper (provided in Appendix~\ref{sec:appendix_emotion_definitions}), for better comparability.

The choice of an LLM fell on the Mistral model\footnote{\href{https://huggingface.co/mistralai/Mistral-7B-Instruct-v0.2}{mistralai/Mistral-7B-Instruct-v0.2}. Details on prompting are in Appendix~\ref{sec:appendix_details}. We also tested prompts with GPT-3.5 to keep them generic. Cf. also \nameref{sec:limitations} for the model choice.}, a leading model with a large context window that allows for the reuse of its outcomes and can be cost-time-effectively used for massive inference \citep{jiang2023mistral}. As the model sometimes hallucinates and generates emotions out of the provided list, we perform manual mapping of frequent labels to our categories (e.g., \textit{anxiety} to \textit{nervousness}, \textit{indignation} to \textit{anger}, \textit{hope} to \textit{optimism}, \textit{happiness} to \textit{joy}), while removing labels for which we did not find a close match (such as \textit{calm} and \textit{focus}). We also store the model's reasoning with emotion explications.

\begin{figure}[t]
  \includegraphics[width=\columnwidth]{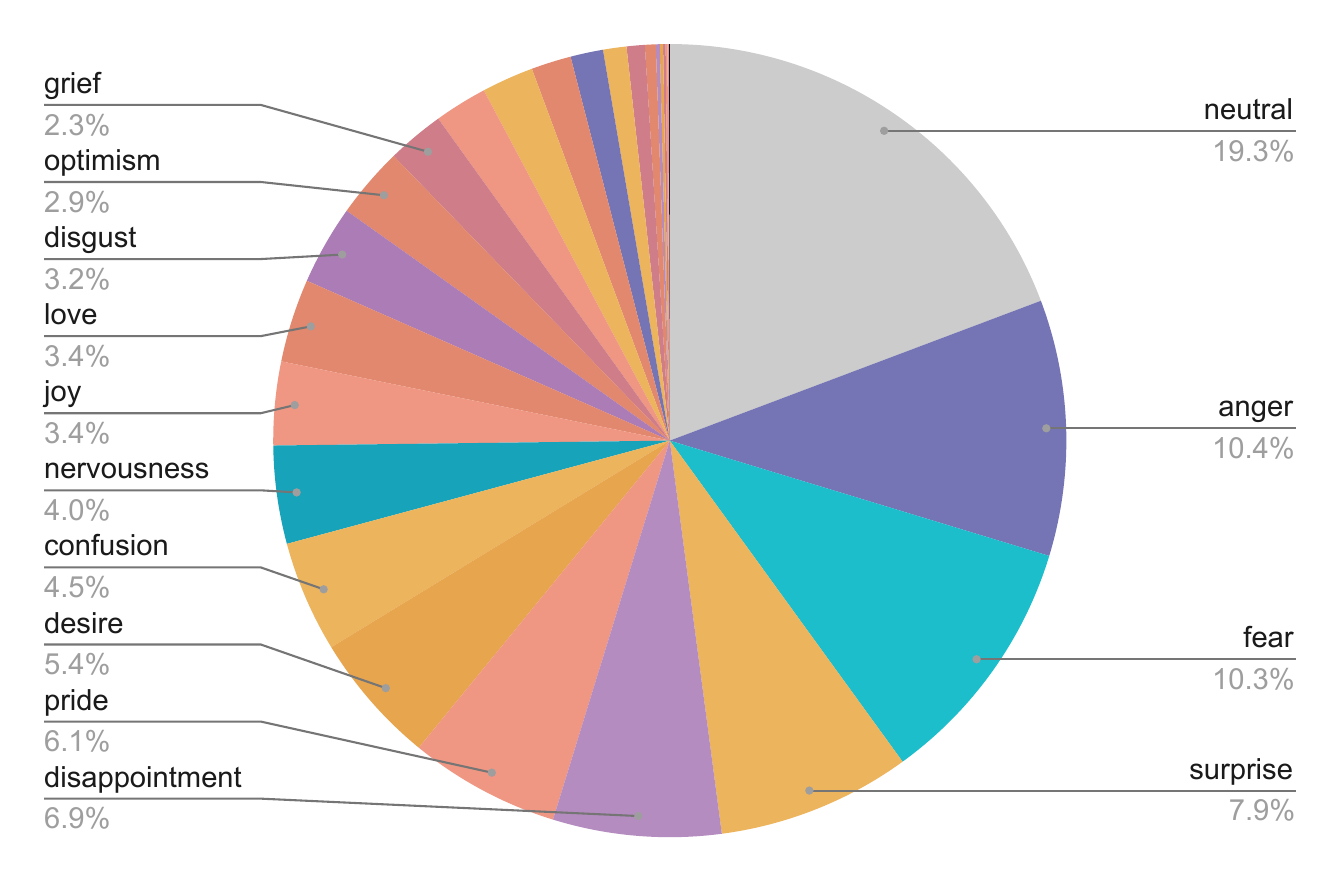}
  \caption{Distribution of primary emotions.}
  \label{fig:primary_distribution}
\end{figure}

\begin{figure}[t]
  \includegraphics[width=\columnwidth]{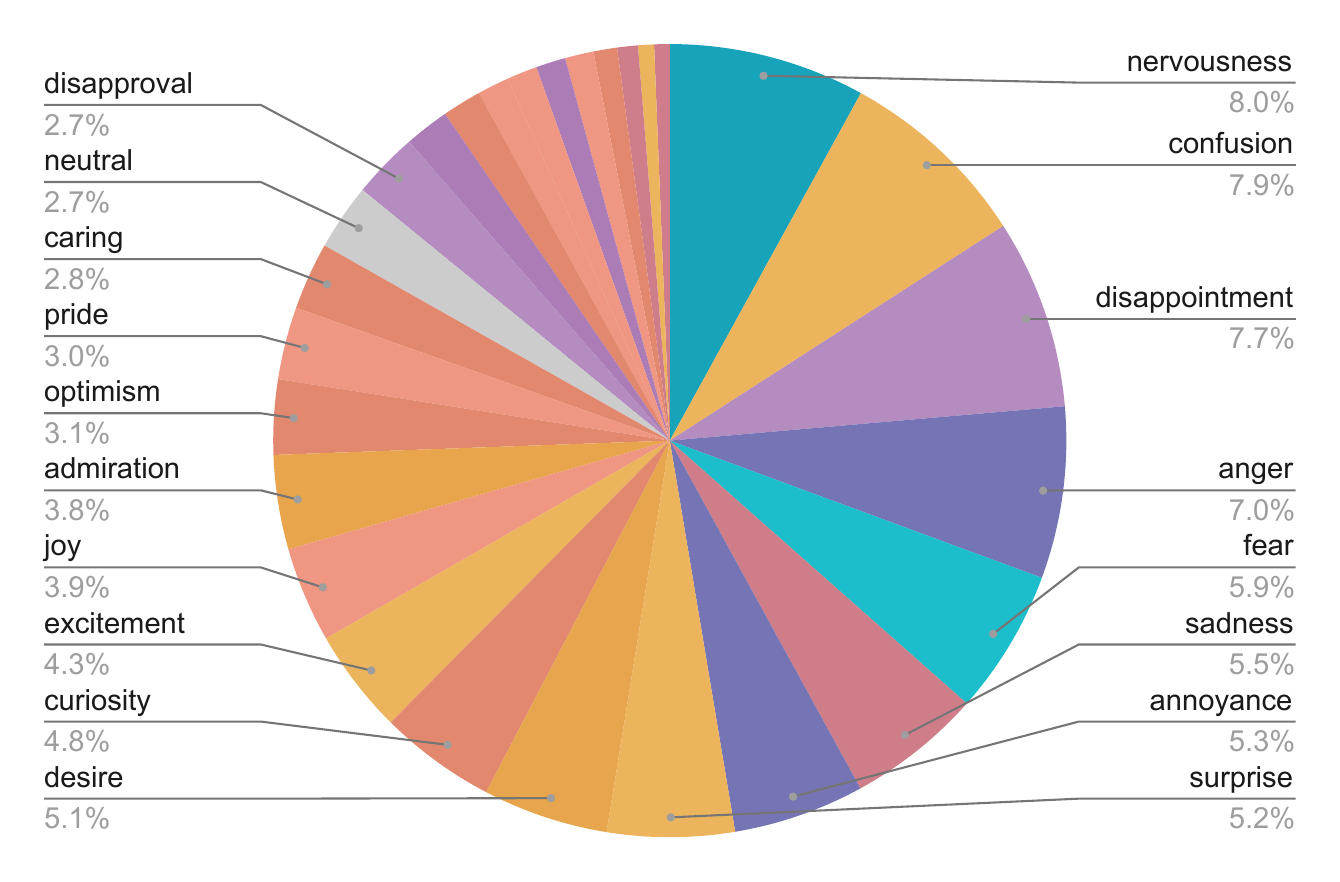}
  \caption{Distribution of soft emotional labels in the dataset (after filtering by the expressiveness level).}
  \label{fig:soft_distribution}
\end{figure}

\begin{figure}
  \includegraphics[width=\columnwidth]{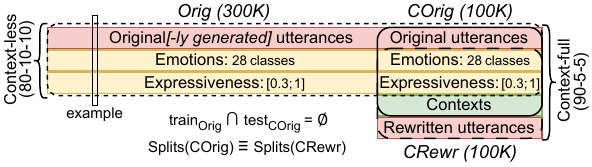}
  \caption{Dataset splits. $Orig$ -- context-less examples, $COrig$ -- context-full examples, $CRewr$ -- the same context-full examples with rewritten utterances.}
  \label{fig:dataset_splits}
\end{figure}

We ran our data synthesis pipeline for 450 GPU hours of NVIDIA H100 iterating over 2000 plots to generate 300K examples for a context-less dataset, a third of which was used for the final pipeline steps, resulting in a context-full dataset of 100K examples (cf. the note on a carbon footprint and scalability in Appendix~\ref{sec:scalability_and_carbon}). On average, 15 actors were extracted from a plot ($std$ of 9.44). The distribution of 300K primary emotions is shown in Figure~\ref{fig:primary_distribution}, while Figure~\ref{fig:soft_distribution} shows the improved class balance after extending them to 1M soft labels for which the expressiveness level was of at least 0.3 (1 to 5 labels per example, with a \textit{mean} of 3.17 and \textit{std} of 0.97). We create 80-10-10 and 90-5-5\footnote{The second dataset is three times smaller. We sampled 90\% for the training to have more examples in absolute value.} splits for the train, dev, and test sets as shown in Figure~\ref{fig:dataset_splits}.

\section{Experiments}
We first perform regular sentiment classification when only an utterance is provided as input. With this, we check that the LLM managed to include emotion-related information in the \textit{original} utterances enough for smaller models to grasp the nuances in various categories and perform the task well. Then, we run experiments using generated contexts paired with either \textit{original} or \textit{rewritten} utterances to investigate whether our contexts, and hence the models trained using them, bring added value for contextual emotion detection tasks. All the base models were trained for multi-label classification with a sigmoid activation function for each category\footnote{Cf. Appendix~\ref{sec:multilabel_eval} for lower boundary choice at evaluation.} and binary cross-entropy loss\footnote{\href{https://pytorch.org/docs/stable/generated/torch.nn.BCELoss.html}{https://pytorch.org/docs/.../torch.nn.BCELoss.html}} using the \textit{transformers} library with the default AdamW optimizer, and initial $learning\_rate$ of 2e-5. We did not activate early stopping and worked with the final checkpoints for all the tasks. The details of their domain fine-tuning are provided in the task-related subsections. The scores are macro-averaged as micro-averaging over-promise results in all tasks.

\subsection{Context-less emotion recognition}
\subsubsection{Emo Pillars synthetic test sets}
\label{subsec:synthetic_context_less}
Our first experiment was directed to study to what extent encoder-decoder transformer models may follow the labels derived using Mistral's inference. We trained two RoBERTa models\footnote{\href{https://huggingface.co/FacebookAI/roberta-large}{FacebookAI/roberta-large}} for sequence classification, one on \textit{original} utterances from the context-less dataset and another on \textit{rewritten} ones from the context-full dataset, excluding contexts. We set $num\_train\_epochs$ to 10, $max\_seq\_length$ to 128, and $batch\_size$ to 64. To contrast the utterance types, we evaluate each model on both of them. We also trained one version of BERT\footnote{\href{https://huggingface.co/google-bert/bert-large-uncased}{google-bert/bert-large-uncased}, used in~\ref{subsec:goemotions}.} \citep{devlin-etal-2019-bert} and Sentence-BERT\footnote{\href{https://huggingface.co/sentence-transformers/paraphrase-distilroberta-base-v1}{sentence-transformers/paraphrase-distilroberta}, in~\ref{subsec:iemocap}.} (SBERT) \citep{reimers-gurevych-2019-sentence} to further fine-tune them on downstream tasks and compare with previous-work models of the same pre-trained architectures. The results are in Table~\ref{tab:results_internal_context_less}.

\begin{table}
  \centering
  \begin{adjustbox}{width=\columnwidth}
  \begin{tabular}{llccc}
    \hline
    \textbf{Model} & \textbf{Eval set} & \textbf{$P$} & \textbf{$R$} & \textbf{$F_1$} \\
    \hline
    \emopitable{emo}{$\pi$}-BERT\textsubscript{\textit{Orig}} & test\textsubscript{\textit{Orig}}     & 0.81 & 0.79 & 0.80  \Tstrut\\
    \emopitable{emo}{$\pi$}-SBERT\textsubscript{\textit{Orig}} & test\textsubscript{\textit{Orig}}    & 0.78 & 0.80 & 0.79  \\
    \emopitable{emo}{$\pi$}-RoBERTa\textsubscript{\textit{Orig}} & test\textsubscript{\textit{Orig}}     & 0.82 & 0.79 & \textbf{0.81}  \\
    \emopitable{emo}{$\pi$}-RoBERTa\textsubscript{\textit{Rewr}} & test\textsubscript{\textit{Orig}}     & 0.70 & 0.78 & 0.74  \\
    \hline
    \emopitable{emo}{$\pi$}-RoBERTa\textsubscript{\textit{Orig}} & test\textsubscript{\textit{COrig}}     & 0.82 & 0.80 & \textbf{0.81}  \Tstrut\\
    \emopitable{emo}{$\pi$}-RoBERTa\textsubscript{\textit{Rewr}} & test\textsubscript{\textit{COrig}}     & 0.71 & 0.78 & 0.74  \\
    \emopitable{emo}{$\pi$}-RoBERTa\textsubscript{\textit{Orig}} & test\textsubscript{\textit{CRewr}}     & 0.68 & 0.63 & 0.65  \\
    \emopitable{emo}{$\pi$}-RoBERTa\textsubscript{\textit{Rewr}} & test\textsubscript{\textit{CRewr}}     & 0.73 & 0.66 & \textbf{0.69}  \\
    \hline
  \end{tabular}
  \end{adjustbox}
  \caption{Context-less intra-dataset evaluation. Subscripts show the type of utterances, while ``\textit{C}'' signifies examples of the context-full test set (contexts excluded).}
  \label{tab:results_internal_context_less}
\end{table}

We observe that the models trained on non-restricted \textit{original} utterances reach high scores when applied to test examples from the same set ($F_1$ of 0.79-0.81). This indicates that there are enough signals in the generated texts to identify emotions even without context. However, such models perform poorer on less emotive material (i.e., \textit{rewritten} utterances) than the model trained for this purpose ($F_1$ of 0.65 vs.\  0.69). On the other hand, the latter model performs worse on longer texts ($F_1$ of 0.74 vs.\  0.81) as it learns to predict emotions beyond explicitly mentioned, and this results in over-labelling and hence low precision.

\subsubsection{GoEmotions}
\label{subsec:goemotions}
GoEmotions is a multi-label categorical dataset of 58K English Reddit comments with 28 emotion labels, including \textit{neutral} class \citep{demszky2020goemotions}. There are some drawbacks in its quality such as rather low inter-annotator agreement and large disparity in terms of emotion frequencies \citep{park2021dimensional}. However, it is still a very valuable source for preliminary evaluation to get a reference point in highly fine-grained emotion classification. We use the original pre-splits for the train, validation, and test sets of the dataset.

We show the performance of the base models applied as is in the upper part of Table~\ref{tab:results_goemotions}, while the bottom part provides models fine-tuned specifically on this dataset. We ran fine-tuning of our pre-trained \emopiintext{emo}{$\pi$} models for 3 epochs with a batch size of 16. Even though our models operate within the same set of categories, fine-tuning allows adapting them to the domain, i.e., to the writing style pertinent to Reddit and specific topics. To the best of our knowledge, the $F_1$-score of 0.55 (\textit{std} of 0.007 over 3 runs) makes our model SOTA on this task\footnote{We provide per-class evaluation scores in Appendix~\ref{sec:per_class_goemotions}.}.

Analysing the GoEmotions data, \citet{yang2023context} showed that in the absence of a context, more labels fit utterances in this dataset. With this in mind, we also found many cases when our over-labelling was not necessarily a poor behaviour.

\begin{table}
  \centering
  \begin{adjustbox}{width=\columnwidth}
  \begin{tabular}{lccc}
    \hline
    \textbf{Model} & \textbf{$P$} & \textbf{$R$} & \textbf{$F_1$} \\
    \hline
    GPT4 \citep{wang2024large}     &  0.10 & 0.17 & 0.13  \\
    GPT4 \citep{kok2023intertwining}     &  - & - & 0.22  \\
    \emopitable{emo}{$\pi$}-BERT\textsubscript{\textit{Orig}}     & 0.26 & 0.42 & 0.28  \\
    \emopitable{emo}{$\pi$}-RoBERTa\textsubscript{\textit{Orig}}     & 0.25 & 0.45 & 0.28  \\
    \emopitable{emo}{$\pi$}-RoBERTa\textsubscript{\textit{Rewr}}     & 0.22 & 0.33 & 0.22  \\
    \hline
    BERT-based \citep{demszky2020goemotions}     &  0.40 & 0.63 & 0.46 \\
    BERT-based \citep{alvarez2021uncovering}\footnotemark     &  - & - & 0.48 \\
    RoBERTa-based \citep{cortiz2022exploring}     & - & - & 0.49  \\
    RoBERTa-based\textsuperscript{GPT-3.5} \citep{kok2023intertwining}     & - & - & 0.51  \\ 
    BERT-based\textsuperscript{BART} \citep{wang2024large}\footnotemark     &  0.52 & 0.53 & 0.52  \\
    \emopitable{emo}{$\pi$}-BERT\textsubscript{\textit{Orig}}-fine-tuned       & 0.51 & 0.57 & \underline{0.54}  \\
    \emopitable{emo}{$\pi$}-RoBERTa\textsubscript{\textit{Orig}}-fine-tuned    &  \meanstd{0.53}{±0.007} & \meanstd{0.58}{±0.007} & \meanstd{\textbf{0.55}}{±0.007}  \Bstrut\\
    \hline
  \end{tabular}
  \end{adjustbox}
  \caption{Evaluation on the GoEmotions task. Superscripts show the models used for data augmentation\protect\footnotemark. \ ``±'' -- standard deviation for three fine-tuned models.}
  \label{tab:results_goemotions}
\end{table}
\footnotetext[14]{Also established a challenging 88-class benchmark on a massive self-reports corpus from  \citep{lykousas2019sharing}.}
\footnotetext[15]{Paraphrasing with BART \citep{lewis2019bart} was used.}
\footnotetext[16]{We do not include models of \citep{park2021dimensional} and \citep{zanwar2022improving} as they used only 7 out of 28 emotion labels.}

\subsubsection{ISEAR}
\label{subsec:isear}
ISEAR (International Survey on Emotion Antecedents
and Reactions) is a single-label first-party annotated corpus that contains 7,666 self-reported emotional events within 7 categories \citep{scherer1994evidence}\footnote{Some reports are dummy like \textit{``[No response.]''}, and therefore the total number of examples used is 150 less.}. Similarly to \citet{zanwar2022improving}, we fine-tune and evaluate the model 5 times within 5-fold cross-validation using
an 80/20 split. We also use the same parameter values: 8 epochs, $batch\_size$ of 4 and $max\_seq\_length$ of 512. We keep the size of the outcome layer as 28 to encourage updating the model weights only for the sake of domain adaptation, without a need to refit them to the low-dimensional outcome. The same as \citet{yang2023context}, we map emotions of \textit{shame} and \textit{guilt} to \textit{embarrassment} and \textit{remorse} correspondingly, while emotions of \textit{anger}, \textit{disgust}, \textit{fear}, \textit{sadness}, and \textit{joy} are mapped 1-to-1. Should the fine-tuned multi-label model predict optional emotions beyond the task, they are ignored in the evaluation. We also approach SOTA on this task (cf. Table~\ref{tab:results_isear}). As for non-fine-tuned modes, BERT trained on the GPT-4-based extension of the GoEmotions dataset \citep{yang2023context} performs poorer than our BERT.

\begin{table}
  \centering
  \begin{adjustbox}{width=\columnwidth}
  \begin{tabular}{lccc}
    \hline
    \textbf{Model} & \textbf{$P$} & \textbf{$R$} & \textbf{$F_1$} \\
    \hline
    GoEmotions\textsuperscript{GPT-4}-BERT \citep{yang2023context}     & - & - & 0.31  \Tstrut\\
    \emopitable{emo}{$\pi$}-BERT\textsubscript{\textit{Orig}}     & 0.61 & 0.58 & 0.56  \Tstrut\\
    \emopitable{emo}{$\pi$}-RoBERTa\textsubscript{\textit{Orig}}     & 0.62 & 0.65 & 0.61  \Tstrut\\
    \emopitable{emo}{$\pi$}-RoBERTa\textsubscript{\textit{Rewr}}     & 0.62 & 0.62 & 0.59  \Tstrut\\
    \hline
    RoBERTa-based \citep{zanwar2022improving}\footnotemark & - & - & 0.73 \\
    RoBERTa-based \citep{park2021dimensional} & - & - & \textbf{0.75} \\
    \emopitable{emo}{$\pi$}-RoBERTa\textsubscript{\textit{Orig}}-fine-tuned  & \meanstd{\textbf{0.76}}{±0.011} & \meanstd{\textbf{0.75}}{±0.011} & \meanstd{\textbf{0.75}}{±0.013}  \\
    \hline
  \end{tabular}
  \end{adjustbox}
  \caption{Evaluation on the ISEAR task. Superscript shows the model used for text extension in data points. ``±'' -- standard deviation for three fine-tuned models.}
  \label{tab:results_isear}
\end{table}

\footnotetext[18]{Evaluated by the authors only on 5 out of 7 classes.}

\subsubsection{IEMOCAP}
\label{subsec:iemocap}
IEMOCAP (Interactive Emotional Dyadic Motion Capture) is an audio-video-text dataset collected following theatrical theory in order to simulate natural dyadic interactions between actors \citep{busso2008iemocap}. For this task, we do not attach a decoder but plug in the derived textual embeddings into the multi-modal architecture that performs the best on this task, namely CORECT \citep{nguyen-etal-2023-conversation}. CORECT starts from acoustic, visual and textual embeddings derived from pre-trained models and leverages them within a relational temporal graph neural network combined with a pairwise cross-modal feature interaction component.

\begin{table}
  \centering
  \begin{adjustbox}{width=\columnwidth}
  \begin{tabular}{lccc}
    \hline
    \textbf{Model} & \textbf{$P$} & \textbf{$R$} & \textbf{$F_1$} \\
    \hline
    CORECT\textsubscript{4} $\leftarrow$ SBERT \citep{nguyen-etal-2023-conversation}\footnotemark & 0.81 & 0.83 & 0.82 \Tstrut\\
    CORECT\textsubscript{4} $\leftarrow$ \emopitable{emo}{$\pi$}-SBERT\textsubscript{\textit{Orig}}    & 0.86 & 0.78 & 0.81  \Tstrut\\
    CORECT\textsubscript{4} $\leftarrow$ \emopitable{emo}{$\pi$}-RoBERTa\textsubscript{\textit{Orig}}     & 0.82 & 0.84 & \textbf{0.83}  \Tstrut\\
    \hline
    CORECT\textsubscript{6} $\leftarrow$ SBERT \citep{nguyen-etal-2023-conversation} & 0.69 & 0.68 & \textbf{0.67} \\
    CORECT\textsubscript{6} $\leftarrow$ \emopitable{emo}{$\pi$}-SBERT\textsubscript{\textit{Orig}}    & 0.65 & 0.65 & 0.65  \\
    CORECT\textsubscript{6} $\leftarrow$ \emopitable{emo}{$\pi$}-RoBERTa\textsubscript{\textit{Orig}}     & 0.64 & 0.64 & 0.63 \\
    \hline
  \end{tabular}
  \end{adjustbox}
  \caption{Evaluation on the IEMOCAP tasks. Subscripts ``4'' and ``6'' mean the size of the model's output layer.}
  \label{tab:results_iemocap}
\end{table}

\footnotetext{We retrained the authors' model to obtain macro scores.}

We perform evaluation in this multi-modal setup to probe the text representations that our models learnt (encoded in the last-layer hidden state of the first token of the sequence, i.e., classification token). In literature, two IEMOCAP settings are used for testing, one with 4 categories (\textit{anger}, \textit{sadness}, \textit{happiness}, and \textit{neutral}) and one with 6 (adding \textit{excited} and \textit{frustrated}). We experiment with both settings. The multi-modal model is trained with the default parameters and data splits specified in the CORECT's repository\footnote{\href{https://github.com/leson502/CORECT_EMNLP2023/}{https://github.com/leson502/CORECT\_EMNLP2023/}}\footnote{The default SBERT model in CORECT is the same as the one we used for fine-tuning with our synthetic dataset in~\ref{subsec:synthetic_context_less}.}.

The scores are shown in Table~\ref{tab:results_iemocap}. The model trained with our RoBERTa-based embeddings overcomes the SOTA performance in a 4-way task while 

\begin{table}
  \centering
  \begin{adjustbox}{width=\columnwidth}
  \begin{tabular}{llccc}
    \hline
    \textbf{Model} & \textbf{Eval set} & \textbf{$P$} & \textbf{$R$} & \textbf{$F_1$} \\
    \hline
    \emopitable{emo}{$\pi$}-CRoBERTa\textsubscript{\textit{COrig}}\footnotemark & test\textsubscript{\textit{COrig}}     & 0.81 & 0.78 & \textbf{0.79} $\downarrow$\ \textsuperscript{(***)} \Tstrut\\
    \emopitable{emo}{$\pi$}-CRoBERTa\textsubscript{\textit{CRewr}} & test\textsubscript{\textit{COrig}}     & 0.73 & 0.78 & 0.75 $\uparrow$\ \textsuperscript{(***)} \Tstrut\\
    \emopitable{emo}{$\pi$}-CRoBERTa\textsubscript{\textit{COrig}} & test\textsubscript{\textit{CRewr}}     & 0.69 & 0.66 & 0.67 $\uparrow$\ \textsuperscript{(***)} \Tstrut\\
    \emopitable{emo}{$\pi$}-CRoBERTa\textsubscript{\textit{CRewr}} & test\textsubscript{\textit{CRewr}}     & 0.75 & 0.69 & \textbf{0.72} $\uparrow$\ \textsuperscript{(***)} \Tstrut\\
    \hline
  \end{tabular}
  \end{adjustbox}
  \caption{Context-aware intra-dataset evaluation. Arrows indicate the change in corresponding values in Table~\ref{tab:results_internal_context_less}\protect\footnotemark.}
  \label{tab:results_internal_context_aware}
\end{table}

\footnotetext[22]{``\textit{C}'' in model name and input type stands for ``context''.}

\footnotetext[23]{All the gains have statistical significance at the 1\% level according to t-test and Mann-Whitney U Test.}

\noindent also reaching promising figures in a 6-way task. The drop is mainly due to frequent confusion between \textit{excited}, \textit{happiness}, and \textit{neutral} categories\footnote{Table~\ref{tab:confusion_iemocap} in Appendix~\ref{sec:iemocap_confusion} provides a confusion matrix.}.

\subsection{Context-aware emotion recognition}

We evaluate context-aware models within the same lines: first on our dataset and then on a public task.

\subsubsection{Emo Pillars synthetic test sets}

For this setting with large input vectors, we trained only RoBERTa models to save the compute. We increased $max\_seq\_length$ to 512 and reduced $batch\_size$ to 32. Other parameter values remained the same\footnote{We also trained models with varying token types for contexts and utterances, but it didn't affect the scores on the sets.}. The training set size is 90K.

The results of the intra-dataset evaluation are in Table~\ref{tab:results_internal_context_aware}. The main finding is that both models trained with contexts improve scores by 2-3 p{.}p{.} on \textit{rewritten} utterances paired with contexts (cf. the last two lines in Table~\ref{tab:results_internal_context_less}). This confirms that the generated contexts contribute to the clarification of the emotions ambiguously expressed in the utterances. We also see that \mbox{\emopitable{emo}{$\pi$}{-CRoBERTa}\textsubscript{\textit{CRewr}}}, which was encouraged to pay more attention to the context, also improved by 3 p.p. in precision on the less-ambiguous inputs. On the contrary, \mbox{\emopitable{emo}{$\pi$}-CRoBERTa\textsubscript{\textit{COrig}}} performs worse but only by 1 p.p. than its context-less version in both precision and recall on such utterances, meaning that the choice can be safely made towards context-aware models when the context is available.

\begin{table}
  \centering
  \begin{adjustbox}{width=\columnwidth}
  \begin{tabular}{lcccc}
    \hline
    \textbf{Model} & \textbf{Set} & \textbf{$P$} & \textbf{$R$} & \textbf{$F_1$} \\
    \hline
    \emopitable{emo}{$\pi$}-CRoBERTa\textsubscript{\textit{COrig}}  & dev\textsubscript{4}  &  0.60 & 0.51 & 0.53  \Tstrut\\
    \emopitable{emo}{$\pi$}-CRoBERTa\textsubscript{\textit{CRewr}}  & dev\textsubscript{4}   & 0.50 & 0.58 & 0.54  \Tstrut\\
    \hline
     The 1\textsuperscript{st} at \citep{chatterjee2019semeval}  & test\textsubscript{3} & 0.8086 & 0.7873	& 0.7963 \\
    ANA (the 5\textsuperscript{th}) \citep{huang2019ana}\footnotemark  & test\textsubscript{3} & 0.7785 & 0.7713 & 0.7729 \\
    \emopitable{emo}{$\pi$}-CRoBERTa\textsubscript{\textit{COrig}}-fine-tuned  & dev\textsubscript{3} & 0.7467 & 0.7633 & 0.7567 \\
    \emopitable{emo}{$\pi$}-CRoBERTa\textsubscript{\textit{CRewr}}-fine-tuned  & dev\textsubscript{3} & 0.7633 & 0.7833 & 0.7733 \\
    \hline
    \emopitable{emo}{$\pi$}-CRoBERTa\textsubscript{\textit{COrig}}-fine-tuned  & dev\textsubscript{4}  & 0.80 & 0.81 & 0.81  \Tstrut\\
    \emopitable{emo}{$\pi$}-CRoBERTa\textsubscript{\textit{CRewr}}-fine-tuned  & dev\textsubscript{4} & 0.81 & 0.83 & 0.82  \Tstrut\\
    \hline
  \end{tabular}
  \end{adjustbox}
  \caption{Evaluation on the EmoContext task. The number of used classes is in the subscripts of the sets.}
  \label{tab:results_emocontext}
\end{table}

\footnotetext[26]{ANA \citep{huang2019ana} is a BERT-based model. The 1\textsuperscript{st} team did not submit the system description paper.}

\subsubsection{EmoContext}
EmoContext (Contextual Emotion Detection in Text) -- a corpus of dialogues created for the SemEval-2019 contest from user interactions with a conversational agent, each consisting of an emotion-annotated user's utterance provided with two previous dialogue turns as a context \citep{chatterjee2019semeval}. It is a single-label task with four classes -- \textit{happy}, \textit{sad}, \textit{angry}, and \textit{others}. As with ISEAR, we solve it as a multi-label task, making all 28 classes available. The difference, though, is in the absence of direct mapping of the ``\textit{others}'' category to our classes. To avoid forcing our models to learn this new compositional category, we relabelled the ``\textit{others}'' examples in the training set by choosing the most probable label predicted by \mbox{\emopitable{emo}{$\pi$}-CRoBERTa\textsubscript{\textit{COrig}}}\footnote{If the maximum sigmoid score across all classes was lower than 0.3, we assigned the default ``\textit{neutral}'' label.}. We fine-tuned our contextual models on the updated set for 3 epochs.

Table~\ref{tab:results_emocontext} shows the results. As intended, the model trained on \textit{rewritten} utterances (with their contexts) was scored higher than another one trained on more informative utterances, without focusing on the context. We used an open \textit{dev} set for evaluation; the scores of the models ranked the 1\textsuperscript{st} and the 5\textsuperscript{th} on the \textit{test} set at the contest are given for reference\footnote{In \citet{chatterjee2019semeval}, the ``\textit{others}'' class scores excluded from averaging (test$_3$). We do the same for \textit{dev} (dev$_3$).}. The main advantage of the proposed fine-tuning is that our models can predict relevant emotions beyond the restricted taxonomy (mapped to \textit{others} at evaluation).

\begin{table}[t]
  \centering
  \begin{adjustbox}{width=\columnwidth}
  \begin{tabular}{lccc}
    \hline
    \textbf{Model} & \textbf{$P$} & \textbf{$R$} & \textbf{$F_1$} \\
    \hline
    \emopitable{emo}{$\pi$}-CRoBERTa\textsubscript{\textit{CRewr}}-fine-tuned, w/o relab & 0.80 & 0.82 & 0.81  \Tstrut\\
    \emopitable{emo}{$\pi$}-CRoBERTa\textsubscript{\textit{CRewr}}-fine-tuned, w/ relab & 0.81 & \textbf{0.83} & \textbf{0.82}  \Tstrut\\
    \emopitable{emo}{$\pi$}-CRoBERTa\textsubscript{\textit{CRewr}}-fine-tuned, w/ relab, typed & \textbf{0.82} & \textbf{0.83} & \textbf{0.82}  \Tstrut\\
    \hline
  \end{tabular}
  \end{adjustbox}
  \caption{Effect of relabelling the training set and introducing the token types in an extra input vector.}
  \label{tab:results_emocontext_ablation}
\end{table}

\begin{table}
  \centering
  \begin{adjustbox}{width=\columnwidth}
  \begin{tabular}{lcc}
    \hline
    \textbf{Example} & \textbf{GT label} & \textbf{Our label(s)}\\ \hline
    \makecell[{{p{0.61\columnwidth}}}]{User: "Sorry"\\
CA: "Sorry for what??"\\
User: "Don't know what to do"} & 
    \cellcolor{sadness}
    sad &
    \cellcolor{surprise}
    confusion (0.97)\\
    \hline
    \makecell[{{p{0.61\columnwidth}}}]{User: "U R just repeating the old stories"\\
CA: "You're getting old."\\
User: "I am fed up"} & \cellcolor{sadness}
    sad &
    \cellcolor{anger}
    \begin{tabular}{c}annoyance (0.48) \\anger (0.2)\end{tabular}\\
    \hline
    \makecell[{{p{0.61\columnwidth}}}]{User: "But..."\\
CA: "then"\\
User: "I'm feeling nervous"} & 
    \cellcolor{sadness}
    sad &
    \cellcolor{fear}
    \begin{tabular}{c}nervousness (0.99) \\sadness (0.38)\end{tabular}\\
    \hline
  \end{tabular}
  \end{adjustbox}
  \caption{EmoContext examples with improved labels. ``GT'' -- ``ground truth''; ``CA'' -- ``conversational agent''. In parentheses: per-class values of a sigmoid function.}
  \label{tab:error_analysis_emocontext}
\end{table}

Table~\ref{tab:results_emocontext_ablation} provides results of an ablation study that verifies that the models benefit from the training set relabelling and perform similarly to when cross-attention is informed with unique values assigned to tokens of contexts (0s) and utterances (1s) in the \textit{type} vector. We also looked at predictions and found out that our labels sometimes fit better than ground truth (Table~\ref{tab:error_analysis_emocontext}). The problem seems to be in human annotation within a very limited emotion taxonomy, which leads to overgeneralization. 

\section{Use Case}
\label{sec:use_case}
In order to validate our models in a real-world scenario, we applied them to comments on music performances from YouTube. It is of high importance for creators of novel sound experiences to receive timely feedback from their audiences, including a virtual one. Moreover, they need a detailed view of people's perceptions to verify their expectations of evoking certain feelings across a wide spectrum (rather than just knowing whether certain aspects were liked or disliked). As comments are often brief, it is important to situate them in context. We find our models highly relevant to this use case.

\begin{figure}
  \includegraphics[width=\columnwidth]{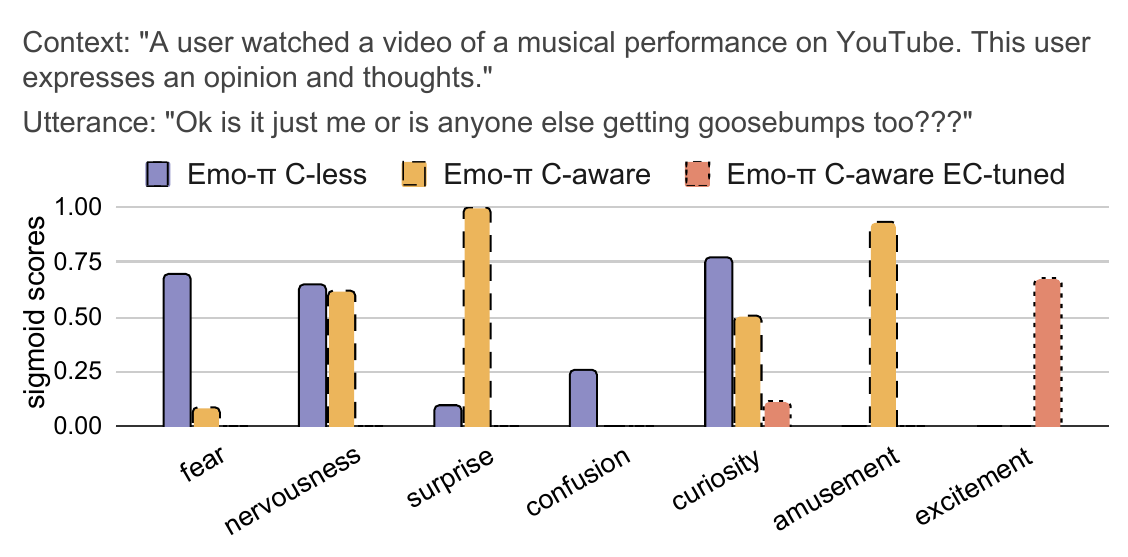}
  \caption{Varied predictions on a YouTube comment.}
  \label{fig:use_case_youtube}
\end{figure}

Figure~\ref{fig:use_case_youtube} contrasts variously trained \emopiintext{emo}{$\pi$} models (the input example is given above the diagram). The context-less model over-interprets the utterance as expressing \textit{fear}. The context-aware model recognizes a variety of more positive emotions, but also with a shade of \textit{nervousness}. The EmoContext-tuned model seems prone to less direct interpretation (\textit{nervousness} is excluded), but it learnt to excessively limit the number of labels. The balance is to be established: for example, the fine-tuning set relabelling might include not only updating ``\textit{others}'' category but also extending the number of labels when the base \emopiintext{emo}{$\pi$} model predicts more of them.

\section{Data analysis}
\label{sec:data_analysis}

To provide insights on the quality of the generated dataset, we encoded contexts and utterances with sentence embeddings\footnote{\href{https://huggingface.co/sentence-transformers/all-mpnet-base-v2}{sentence-transformers/all-mpnet-base-v2}} and made various comparisons within these semantic representations.

\textbf{Semantic diversity of utterances.} We calculated pairwise cosine similarities between the embeddings of 10K random \textit{original} utterances. Low value of $\mu=0.12$ ($\sigma=0.1$, $\eta_{.99}=0.4$) shows that the utterances are rather dissimilar across and within an emotion (cf. also Appendix~\ref{sec:lexical_diversity}). However, we observe reduced diversity among neutral utterances of the same character. Even though they are generated within the same inference, we get $\mu=0.30$ ($\sigma=0.16$, $\eta_{.99}=0.72$); while the similarity between two random neutral utterances in the data is: $\mu=0.16$ ($\sigma=0.12$, $\eta_{.99}=0.52$). Near-duplicates can be filtered out in future training using a high-threshold cut-off over these scores.

\textbf{Emotion coverage.}
As shown in Figure~\ref{fig:soft_distribution}, only a few emotions get a small share in the dataset: similarly to the GoEmotions set, only 8 emotions take less than 1.5\% each. On the other hand, the absolute number of examples per emotion is 50 times larger on average and the balance between positive, negative, and ambiguous polarities\footnote{Cf. Table~\ref{markers-table}; as distributed in \citet{demszky2020goemotions}.} is a little bit better ($33|48|19$ vs. $54|30|16$). \textit{Neutral} class takes 2.7\% and remains only in 9\% examples after filtering by expressiveness.

\textbf{Topic coverage.}
To verify topic diversity, we clustered 3000 utterances within a single class of \textit{joy} to avoid grouping by emotion categories. For this, we formed a graph by creating edges between embeddings with a similarity of more than 0.6 and decomposed it, maximizing the partition modularity \citep{blondel2008fast}. We discovered more than 300 clusters (topics), from which 20 had a large number of utterances. Topics were named using GPT-3.5: they vary from generic to very specific (e.g., ``overcoming adversity or conflict'' or ``achievements in medical procedures''). More details are in Appendix~\ref{sec:appendix_topic_coverage}.

\begin{figure}[t]
  \includegraphics[width=\columnwidth]{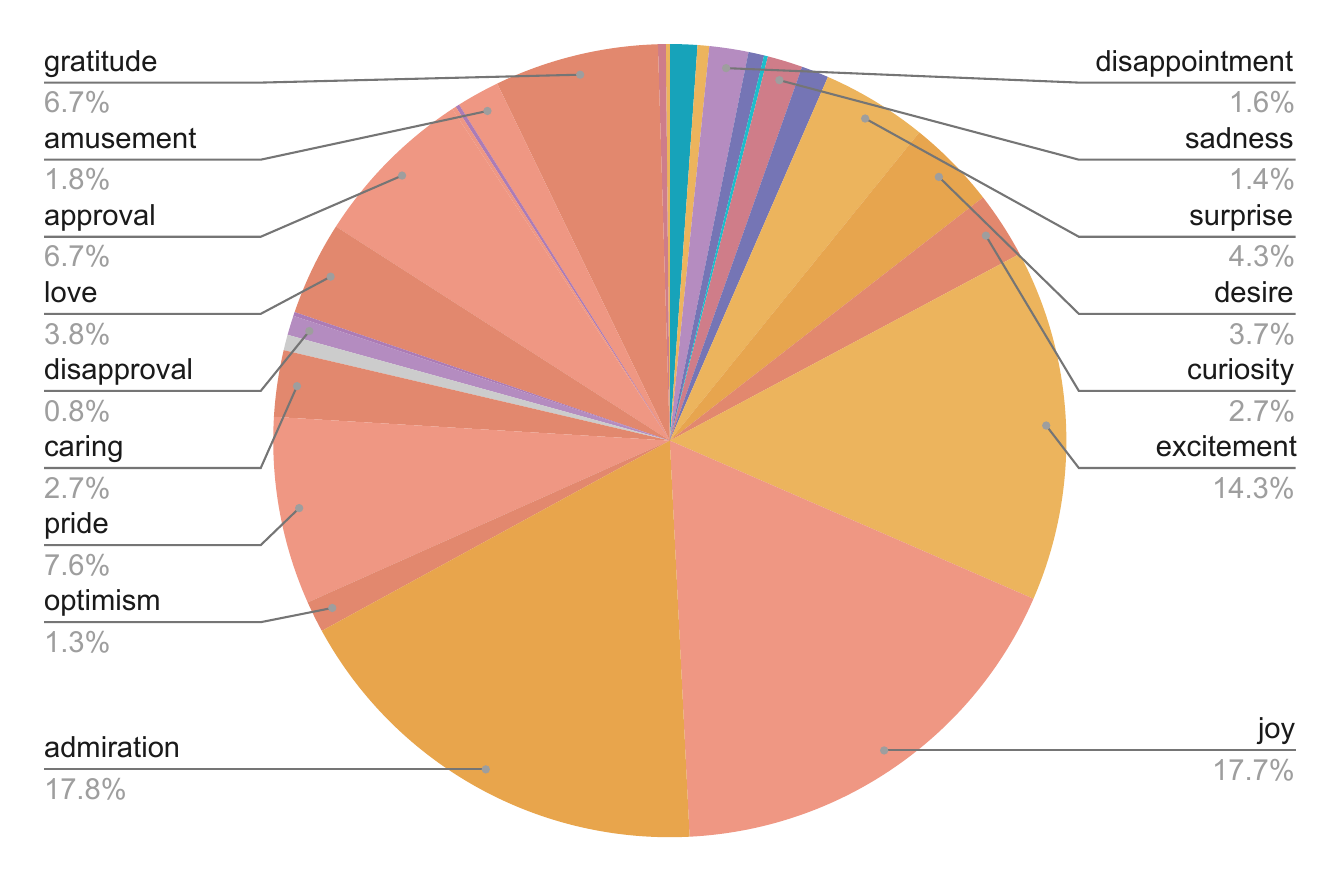}
  \caption{Emotions in the topic of music performances.}
  \label{fig:soft_distribution_music}
\end{figure}

\textbf{Relationship between topics and emotions.} We found a group of 200 examples semantically similar to the data in our use case (i.e. music performances) and looked at the distribution of emotions within the topic. Emotions are very diverse but skewed towards positive (82\% of labels; Figure~\ref{fig:soft_distribution_music}), which greatly differs from the global distribution in Figure~\ref{fig:soft_distribution}. This suggests that polarity generation depends on the nature of the seed plots (their topics) rather than on the LLM's biases.

\textbf{Context personalization.}
Comparing contexts derived from the same plot (iterating over 1000 plots), we confirm that the generated summaries are personalized. The similarity of contexts for the same actor is high: $\mu=0.77$ ($\sigma=0.15$); while across actors of the same story, they are still similar but considerably less: $\mu=0.56$ ($\sigma=0.17$). Contexts across stories are dissimilar: $\mu=0.24$ ($\sigma=0.11$), and six times smaller than their plots (145 tokens, $\sigma=64$ vs. 823 tokens, $\sigma=582$).

\textbf{Changes after cleaning and rewriting.}
The content of the contexts is only slightly changed after cleaning from emotions; the similarity is: $\mu=0.92$ ($\sigma=0.08$, $\eta_{.99}=1.0$; 10K pairs). The \textit{rewritten} utterances are shorter than the \textit{original}: 16 vs.\ 20 tokens on average; the similarity is: $\mu=0.78$ ($\sigma=0.15$, $\eta_{.99}=1.0$; 10K pairs), which means that the meaning is preserved. The style does not deviate much either: 66\% of the utterance ($\sigma=16$) is nearly a substring of the \textit{original}, according to the partial Levenshtein ratio\footnote{\href{https://pypi.org/project/fuzzywuzzy/}{pypi.org/project/fuzzywuzzy/}; it is a max similarity between the shorter string and any same-length part of the longer.}.

\section{Human Evaluation}
\label{sec:human_evaluation}
We carried out a human evaluation of our contextual dataset. Following \citet{sabour-etal-2024-emobench}, we created a multiple-choice task with several plausible choices per example (our set of Mistral-based labels is one of them), a few less plausible choices, and a \textit{none} option (7 in total, mixed). Along, we asked to assess neutrality and suggest missed/unfit emotions\footnote{Entire task setup and discussion are in Appendix~\ref{sec:appendix_human_evaluation}.}. Three postdocs with a degree in computer science annotated 200 examples. The inter-annotator agreement using Cohen’s Kappa
\citep{kohen1960coefficient}, $\kappa = 0.365$ (higher than $\kappa = 0.293$ in \citet{demszky2020goemotions}), points to the high task subjectivity. The accuracy of our labels reaches 0.86 and 0.7 on examples where all three or at least two votes coincide, respectively. Neutrality is perceived too differently, but still gains a high recall of 0.9. The main encountered issue is the incompatibility of some emotions in the soft label sets that originate from the manual mapping provided in Section~\ref{sec:dataset_generation}. Positively, the relevance of contexts and expressiveness rankings was confirmed (cf. Appendix~\ref{sec:appendix_human_evaluation}).

\section{Conclusions}
We proposed a pipeline that generates diverse, labelled synthetic data by extracting knowledge from LLMs for fine-grained context-less and context-aware emotion recognition. We created a voluminous dataset and trained mid-sized language models that show portability to various domains and smaller emotion taxonomies, optionally extending them. Our models also derive sentiment embeddings beneficial for multi-modal setups.

Our approach is to an extent rooted in early works on knowledge extraction from human experts and expert systems that involves retrieving and organizing information from experts to create structured knowledge. We use thought-provoking and clarifying questions in a free-form manner at the start, narrowing in specificity as the pipeline progresses, echoing questionnaires and protocol analysis techniques that allow taking thinking-out-loud as data \citep{olson1987extracting}. We also draw on the idea that only a relatively small amount of an expert's knowledge is potentially relevant in any given situation \citep{mcdermott1983extracting}. By placing an LLM in various story-character situations, we aim to activate different parts of its parametric memory to generate diverse but high-probability token sequences with plausible semantics. 

Future work is concerned with creating data in multiple languages, improving neutral examples, scaling the pipeline for more emotions and other types of text sources, balancing explicitly and implicitly expressed sentiments in the training set, leveraging label explications for aspect-oriented analysis, and applying explainability techniques to get more focused access to the required knowledge.

\pagebreak
\section{Limitations}
\label{sec:limitations}
Firstly, although the provided evaluation gives an idea of the scores our models gained, they may have been held back by the problems in the ground-truth labels that we discuss in the paper.

Secondly, we use a single LLM model to generate the dataset, as creating one of this size is expensive, even with public models like Mistral (requiring 450 GPU hours on top-tier GPUs, such as the NVIDIA H100). Moreover, looking for another well-performing model within our pipeline goes beyond the scope of the paper. However, we hope that the pipeline possesses generalization ability, as we designed prompts using two models (Mistral and GPT-3.5) and obtained reasonable outputs with both of them. On the other hand, sticking to one model allows us to identify more easily whether some problems come from the generated outputs (thus from the chosen model) or from the design of the pipeline or the evaluation task.

Finally, this work relies on LLMs as a backbone for emotion labelling that carries cultural/language implications depending on the data an LLM was pre-trained with (often English-centered or biased towards dominant cultural factors within a language). On the other hand, for domain tuning, we use specific downstream datasets that necessarily introduce further biases in our models. To account for cultural background within a language and across languages, highly multilingual LLMs like Salamandra \citep{gonzalez2025salamandra}, which are trained on carefully curated data with a large variety of sources per language \citep{palomar2024curated, brack2024community} ensuring wide coverage of various cultural aspects, can be a promising alternative to use in the pipeline. A separate effort is required to design domain-specific annotations on a smaller scale for final fine-tuning, depending on the application needs.

\section{Ethical Considerations}
\label{sec:ethical}
The use of various narrative-like text sources as input to the proposed pipeline would require thorough legal research, especially should the texts be related to real people, e.g., news items or posts on social media, since the generation of possible utterances may touch personal interests or lead to potentially harmful texts. Ethics sheet by \citet{mohammad2022ethics} provides further general ethical considerations for automatic emotion recognition and sentiment analysis.

\section{Acknowledgements}
This research was funded by the EC-funded research and innovation programme Horizon Europe under grant agreement number 101070278. Sincere thanks to the three postdoctoral researchers from Pompeu Fabra University and ADAPT Research Centre in Dublin for their invaluable help with the human evaluation.

\bibliography{acl_latex}

\appendix

\section{Emotion Category Definitions}
\label{sec:appendix_emotion_definitions}
\textbf{admiration}: Finding something impressive or worthy of respect\footnote{We removed emoticons from the definitions introduced in \citet{demszky2020goemotions} to keep the style of prompts generic.}.

\noindent\textbf{amusement}: Finding something funny or being entertained.

\noindent\textbf{anger}: A strong feeling of displeasure or antagonism.

\noindent\textbf{annoyance}: Mild anger, irritation.

\noindent\textbf{approval}: Having or expressing a favorable opinion.

\noindent\textbf{caring}: Displaying kindness and concern for others.

\noindent\textbf{confusion}: Lack of understanding, uncertainty.

\noindent\textbf{curiosity}: A strong desire to know or learn something.

\noindent\textbf{desire}: A strong feeling of wanting something or wishing for something to happen.

\noindent\textbf{disappointment}: Sadness or displeasure caused by the nonfulfillment of one’s hopes or expectations.

\noindent\textbf{disapproval}: Having or expressing an unfavorable opinion.

\noindent\textbf{disgust}: Revulsion or strong disapproval aroused by something unpleasant or offensive.

\noindent\textbf{embarrassment}: Self-consciousness, shame, or awkwardness.

\noindent\textbf{excitement}: Feeling of great enthusiasm and eagerness.

\noindent\textbf{fear}: Being afraid or worried.

\noindent\textbf{gratitude}: A feeling of thankfulness and appreciation.

\noindent\textbf{grief}: Intense sorrow, especially caused by someone’s death.

\noindent\textbf{joy}: A feeling of pleasure and happiness.

\noindent\textbf{love}: A strong positive emotion of regard and affection.

\noindent\textbf{nervousness}: Apprehension, worry, anxiety.

\noindent\textbf{optimism}: Hopefulness and confidence about the future or the success of something.

\noindent\textbf{pride}: Pleasure or satisfaction due to ones own achievements or the achievements of those with whom one is closely associated.

\noindent\textbf{realization}: Becoming aware of something.

\noindent\textbf{relief}: Reassurance and relaxation following release from anxiety or distress.

\noindent\textbf{remorse}: Regret or guilty feeling.

\noindent\textbf{sadness}: Emotional pain, sorrow.

\noindent\textbf{surprise}: Feeling astonished, startled by something unexpected.

\noindent\textbf{neutral}: Neutral sentiment.

\section{Additional Details for the Pipeline}
\label{sec:appendix_details}

This section provides details for running inference with the Mistral model in our pipeline and demonstrates its outcomes within the intermediate steps.

We used greedy decoding to increase the plausibility of semantics, reduce the inference time and enable reproducibility; the repetition penalty was set to 1.03; the maximum number of new tokens was set for actor extraction (300), generation of utterances (500), generation of soft labels (100), generation of contexts (300), their cleaning (300), and utterance rewriting (300). To simplify the prompt design process we used the model instance ran in HuggingFaceHub\footnote{\href{https://huggingface.co/docs/hub/en/index}{https://huggingface.co/docs/hub/en/index}}, while the massive generation was done on HPC servers using \textit{transformers} library\footnote{\href{https://github.com/huggingface/transformers/}{https://github.com/huggingface/transformers/}}\footnote{The outcomes in Table~\ref{prompts-table} therefore may slightly differ from the ones generated for the dataset.}.

Table~\ref{prompts-table} shows the designed prompts for each step in the pipeline, along with their corresponding model outcomes.

\section{Compute resources and scalability}
\label{sec:scalability_and_carbon}

\paragraph{Note on a carbon footprint.} The generation of the dataset took us about 400 GPU hours using NVIDIA H100 (200h for the context-less dataset, and 200h for the remaining steps for a three-times smaller context-full dataset). To ensure the proper use of compute, before generating the datasets entirely, we first carried out preliminary experiments on 18K context-full examples (generated in 44h), achieving satisfactory results within the GoEmotions task and intra-dataset evaluation.

\paragraph{Discussion on scalability.} The time it takes to create a dataset for a custom topic depends on the size of the seed corpus to be processed, which should not be large enough to create a valuable dataset. In our work, we used 2000 plots, but even starting with 200, we were able to reach prominent evaluation scores, albeit somewhat lower than those of state-of-the-art models: $0.77$ vs. $0.79$ for RoBERTa on intra-dataset context-aware evaluation and $0.5$ vs. $0.54$ for BERT on GoEmotions. Thus, the substantial decrease in computing with only a small drop in the scores allows for a more accessible realization.

\begin{table*}
  \centering
  \begin{adjustbox}{width=0.8\textwidth}
  \begin{tabular}{|c|c|c|c|c|c|}
    \hline
    \cellcolor{anger}
    \begin{tabular}{c}
        \textbf{anger}  \\
        \textbf{disappointment} \\
        \textbf{annoyance} \\
        \textbf{disapproval} \\
        \textbf{disgust}
    \end{tabular}
    & \cellcolor{sadness}
    \begin{tabular}{c}
        \textbf{sadness} \\
        \textbf{grief} \\
        \textbf{remorse}
    \end{tabular}
    & \cellcolor{fear} 
    \begin{tabular}{c}
        \textbf{fear} \\
        \textbf{nervousness} \\
        \textbf{embarrassment}
    \end{tabular}
    & \cellcolor{surprise}
    \begin{tabular}{c}
        \textbf{surprise} \\
        \textbf{confusion} \\
        \textbf{curiosity} \\
        \textbf{amusement} \\
        \textbf{realization}
    \end{tabular}
    & \cellcolor{optimism}
    \begin{tabular}{c}
        \textbf{optimism} \\
        \textbf{desire} \\
        \textbf{caring}
    \end{tabular}
    & \cellcolor{joy}
   \begin{tabular}{c}
        \textbf{excitement} \\
        \textbf{admiration} \\
        \textbf{joy\ \ pride\ \ love} \\
        \textbf{relief\ \ approval} \\
        \textbf{gratitude}
    \end{tabular}\\
    \hline
\makecell[{{}}]{ horribly (1.0)\\
disorderly (.75)\\
cowardly (.73)\\
wrongly (.38)
} & 
\makecell[{{}}]{ tragically (.50)\\
sorely (.50)\\
terribly (.50)\\
dearly (.21)
} & 
\makecell[{{}}]{ vastly (.80)\\
accidentally (.51)\\
secretly (.43)\\
safely (.29)
} & 
\makecell[{{}}]{ rightly (1.0)\\
usually (.58)\\
unconsciously (.50)\\
magically (.43)
} & 
\makecell[{{}}]{ mutually (1.0)\\
hopefully (1.0)\\
safely (.34)\\
decisively (.31)} & 
\makecell[{{}}]{ brilliantly (1.0)\\
tirelessly (.67)\\
happily (.57)\\
finally (.54)
} \\ 
\hline
  \end{tabular}
  \end{adjustbox}
  \caption{\label{markers-table}
    Markers that reduce emotion ambiguity according to the LLM's ``belief''. In parentheses: the ratio across the groups indicates the strength of association. Each marker is shared by at least two emotions in the group. The three left columns correspond to negative polarity, the two on the right to positive, and the middle to ``ambiguous''.}
\end{table*}

\section{Details on Multi-Label Evaluation}
\label{sec:multilabel_eval}
\begin{figure}[t]
  \includegraphics[width=\columnwidth]{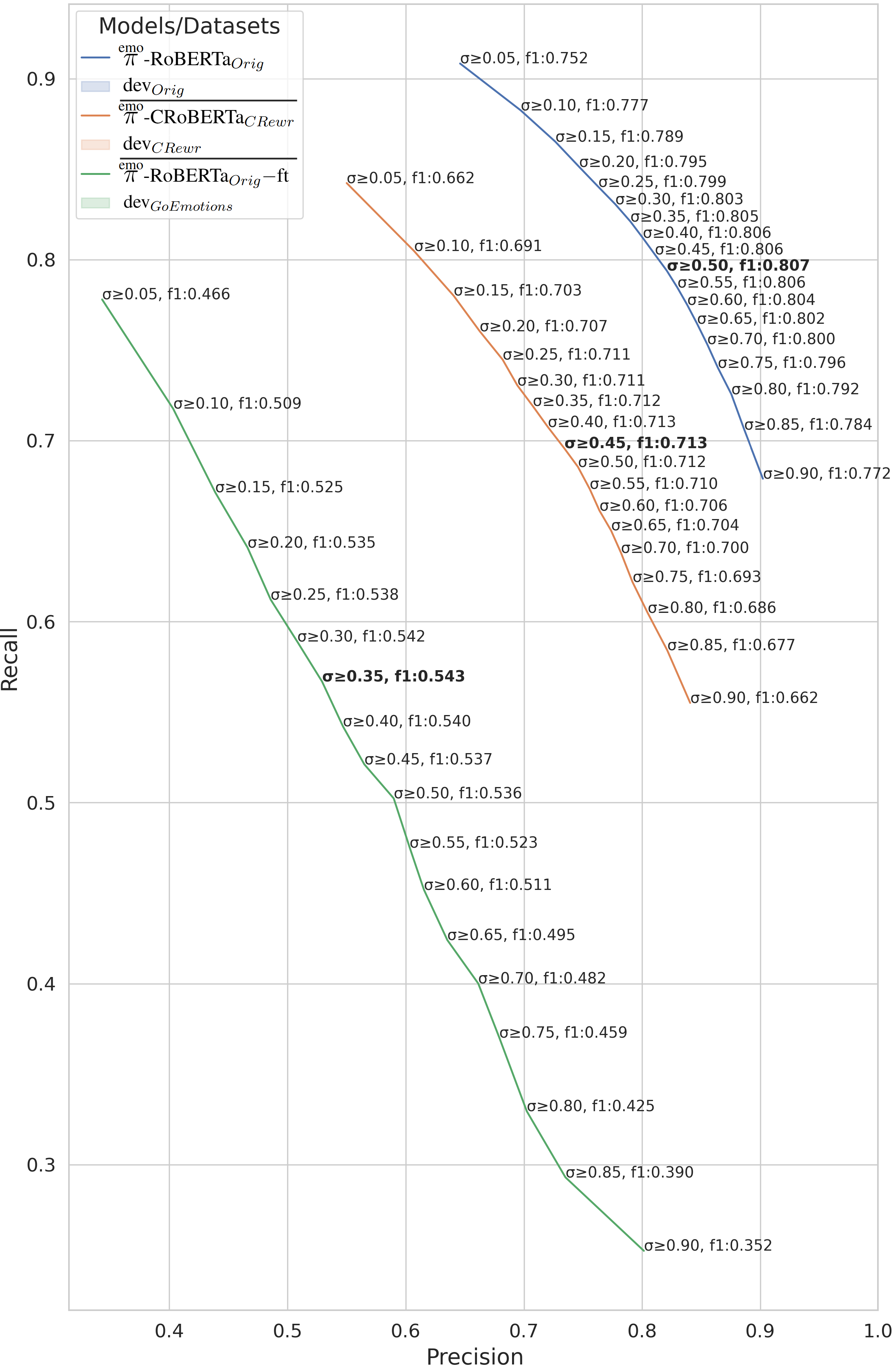}
  \caption{Precision-Recall curves based on different lower boundaries for sigmoid scores (on the dev sets of various tasks; boundaries are equal for all the classes within a task). ``f1'' -- macro $F_1$-score. The largest $F_1$ per task and corresponding boundaries are in bold.}
  \label{fig:sigmas}
\end{figure}

The selection of resulting labels in the model's outcome depends on the lower boundary for predicted sigmoid scores. Using a dev set, we iterate over boundaries from 0.05 to 0.95 with a step of 0.01 and identify one that yields the highest macro $F_1$-score. A single lower value is established for all classes within a task. We show precision-recall curves constructed within these iterations for a context-less and a context-aware model, as well as a model fine-tuned on GoEmotions in Figure~\ref{fig:sigmas}. We can see that the chosen boundaries differ across tasks.

\section{GoEmotions Per-Class Evaluation Scores}
\label{sec:per_class_goemotions}
Table~\ref{tab:class_scores_goemotions} provides per-class evaluation scores and the gains concerning the scores presented in \citet{demszky2020goemotions}.

\begin{table}
  \centering
  \begin{adjustbox}{width=\columnwidth}
  \begin{tabular}{lccccc}
  \hline
  \textbf{Emotion} &	\textbf{$P$} & \textbf{$R$} & \textbf{$F_1$} & \textbf{GoEmotions $F_1$} & \textbf{Gain}\\
  \hline
admiration & 0.64 & 0.80 & \textbf{0.71} & 0.65 & 0.06\\
amusement & 0.75 & 0.93 & \textbf{0.83} & 0.8 & 0.03\\
anger & 0.51 & 0.56 & \textbf{0.53} & 0.47 & 0.06\\
annoyance & 0.36 & 0.40 & \textbf{0.38} & 0.34 & 0.04\\
approval & 0.40 & 0.44 & \textbf{0.42} & 0.36 & 0.06\\
caring & 0.44 & 0.45 & \textbf{0.44} & 0.39 & 0.05\\
confusion & 0.41 & 0.53 & \textbf{0.46} & 0.37 & 0.09\\
curiosity & 0.49 & 0.69 & \textbf{0.57} & 0.54 & 0.03\\
desire & 0.58 & 0.48 & \textbf{0.53} & 0.49 & 0.04\\
disappointment & 0.40 & 0.33 & \textbf{0.36} & 0.28 & 0.08\\
disapproval & 0.43 & 0.48 & \textbf{0.46} & 0.39 & 0.07\\
disgust & 0.47 & 0.49 & \textbf{0.48} & 0.45 & 0.03\\
embarrassment & 0.55 & 0.46 & \textbf{0.50} & 0.43 & 0.07\\
excitement & 0.44 & 0.50 & \textbf{0.47} & 0.34 & 0.13\\
fear & 0.57 & 0.76 & \textbf{0.65} & 0.6 & 0.05\\
gratitude & 0.91 & 0.91 & \textbf{0.91} & 0.86 & 0.05\\
grief & 0.5 & 0.67 & \textbf{0.57} & 0 & 0.57\\
joy & 0.59 & 0.69 & \textbf{0.64} & 0.51 & 0.13\\
love & 0.74 & 0.92 & \textbf{0.82} & 0.78 & 0.04\\
nervousness & 0.44 & 0.48 & \textbf{0.46} & 0.35 & 0.11\\
neutral & 0.57 & 0.59 & 0.58 & 0.68 & -0.10\\
optimism & 0.57 & 0.50 & \textbf{0.53} & 0.51 & 0.02\\
pride & 0.32 & 0.25 & 0.28 & 0.36 & -0.08\\
realization & 0.20 & 0.18 & 0.19 & 0.21 & -0.02\\
relief & 0.56 & 0.82 & \textbf{0.67} & 0.15 & 0.52\\
remorse & 0.54 & 0.62 & 0.58 & 0.66 & -0.08\\
sadness & 0.54 & 0.60 & \textbf{0.57} & 0.49 & 0.08\\
surprise & 0.68 & 0.67 & \textbf{0.68} & 0.5 & 0.18\\
    \hline
Micro average & 0.58 & 0.64 & \textbf{0.61} & - & -\\
    \hline
Macro average & 0.53 & 0.58 & \textbf{0.55} & 0.46 & 0.09\\
STD & 0.14 & 0.19 & 0.16 & 0.19 & 0.14\\
    \hline
  \end{tabular}
  \end{adjustbox}
  \caption{Per-class scores for the GoEmotions task.}
  \label{tab:class_scores_goemotions}
\end{table}

\begin{table}
  \centering
  \begin{adjustbox}{width=0.75\columnwidth}
  \begin{tabular}{lcccccc}
  \hline
  & \textbf{hap}	& \textbf{sad}	& \textbf{neu} &	\textbf{ang} &	\textbf{exc} &	\textbf{fru}\\
  \hline
\textbf{hap} &	74	&4	&\textbf{23}	&0	&\textbf{39}	&4\\
\textbf{sad} & 2	&184	&21	&2	&0	&36\\
\textbf{neu} & \textbf{20}	&17	&259	&19	&\textbf{19}	&50\\
\textbf{ang} &	0	&3	&7	&121	&0	&39\\
\textbf{exc} & \textbf{56}	&1	&\textbf{44}	&6	&187	&5\\
\textbf{fru} & 	1	&16	&72	&57	&2	&233\\
    \hline
  \end{tabular}
  \end{adjustbox}
  \caption{Confusion matrix for the IEMOCAP task.}
  \label{tab:confusion_iemocap}
\end{table}

\section{IEMOCAP Confusion Matrix}
\label{sec:iemocap_confusion}
Table~\ref{tab:confusion_iemocap} provides confusion scores between classes in the IEMOCAP task. We highlight the number of examples confused between \textit{excited}, \textit{happiness}, and \textit{neutral} (201 in total). \textit{Frustration} is confused with \textit{sadness} and \textit{anger} but less (148 in total).

\section{Additional Lexicon Diversity Analysis}
\label{sec:lexical_diversity}
The nature of the dataset allows analyzing how large language models ``transmit'' emotions. In this section, we consider LLM's choices in lexicon.

By subtracting the frequencies of \textit{Rewr} words from the frequencies of \textit{Orig} words, we can get the words that were taken out of utterances by the LLM during rewriting as possibly being suggestive emotional markers. We select a word as a marker if it was removed in at least 60\% of its occurrences in the \textit{Orig} utterances. Then we assign each selected word to a set of emotions with which it appeared more frequently (and at least 5\% of the time with an emotion). Finally, we group emotions that have a large overlap in markers. Table~\ref{markers-table} presents 6 groups and a subset of adverb markers (we focused only on adverbs as they almost do not depend on topics).

We find the groups of emotion categories meaningful, with a clear way of ordering them from negative to positive sentiments, as shown in the header row of the table. This grouping is more fine-grained and informative than the ``positive'', ``negative'', and ``ambiguous'' clustering proposed in \citet{demszky2020goemotions}. The markers are descriptive, and we can see that ``safely'' is indeed related to the emotions of \textit{fear} and \textit{caring} (cf. Section~\ref{subsec:context_importance}).

The large variety of lexicon items in the resource of emotive markers we compiled supports the success of our measures to diversify the content. Furthermore, Table~\ref{tab:sims_emotions_utterances} provides per-class pairwise cosine similarities between utterances in the context-full dataset, showing that diversity is high not only across classes but also within each class.

\begin{table}
  \centering
  \begin{adjustbox}{width=0.7\columnwidth}
  \begin{tabular}{lccr}
  \hline
  \textbf{Main emotion} &	\textbf{$\mu$} & \textbf{$\sigma$} & \textbf{Sample size} \\
  \hline
admiration & 0.20 & 0.13 & 105\\
amusement & 0.18 & 0.09 & 1923\\
anger & 0.20 & 0.11 & 9487\\
annoyance & 0.19 & 0.11 & 1212\\
approval & 0.19 & 0.12 & 75\\
caring & 0.22 & 0.12 & 34\\
confusion & 0.17 & 0.10 & 4124\\
curiosity & 0.16 & 0.12 & 406\\
desire & 0.16 & 0.11 & 4885\\
disappointment & 0.19 & 0.10 & 6344\\
disapproval & 0.18 & 0.11 & 147\\
disgust & 0.20 & 0.11 & 2842\\
embarrassment & 0.28 & 0.12 & 93\\
excitement & 0.22 & 0.10 & 1873\\
fear & 0.23 & 0.13 & 9411\\
gratitude & 0.25 & 0.11 & 1537\\
grief & 0.27 & 0.13 & 2051\\
joy & 0.23 & 0.11 & 3055\\
love & 0.24 & 0.12 & 3108\\
nervousness & 0.21 & 0.11 & 3645\\
optimism & 0.23 & 0.12 & 2723\\
pride & 0.22 & 0.12 & 5514\\
realization & 0.18 & 0.11 & 858\\
relief & 0.29 & 0.12 & 47\\
remorse & 0.45 & 0.01 & 3\\
sadness & 0.24 & 0.12 & 655\\
surprise & 0.17 & 0.10 & 7069\\
neutral & 0.16 & 0.12 & 10000\\
    \hline
  \end{tabular}
  \end{adjustbox}
  \caption{Per-class pairwise utterance similarity scores in the context-full dataset.}
  \label{tab:sims_emotions_utterances}
\end{table}

\section{Topic Analysis Details}
\label{sec:appendix_topic_coverage}

\begin{figure}[!t]
  \includegraphics[width=\columnwidth]{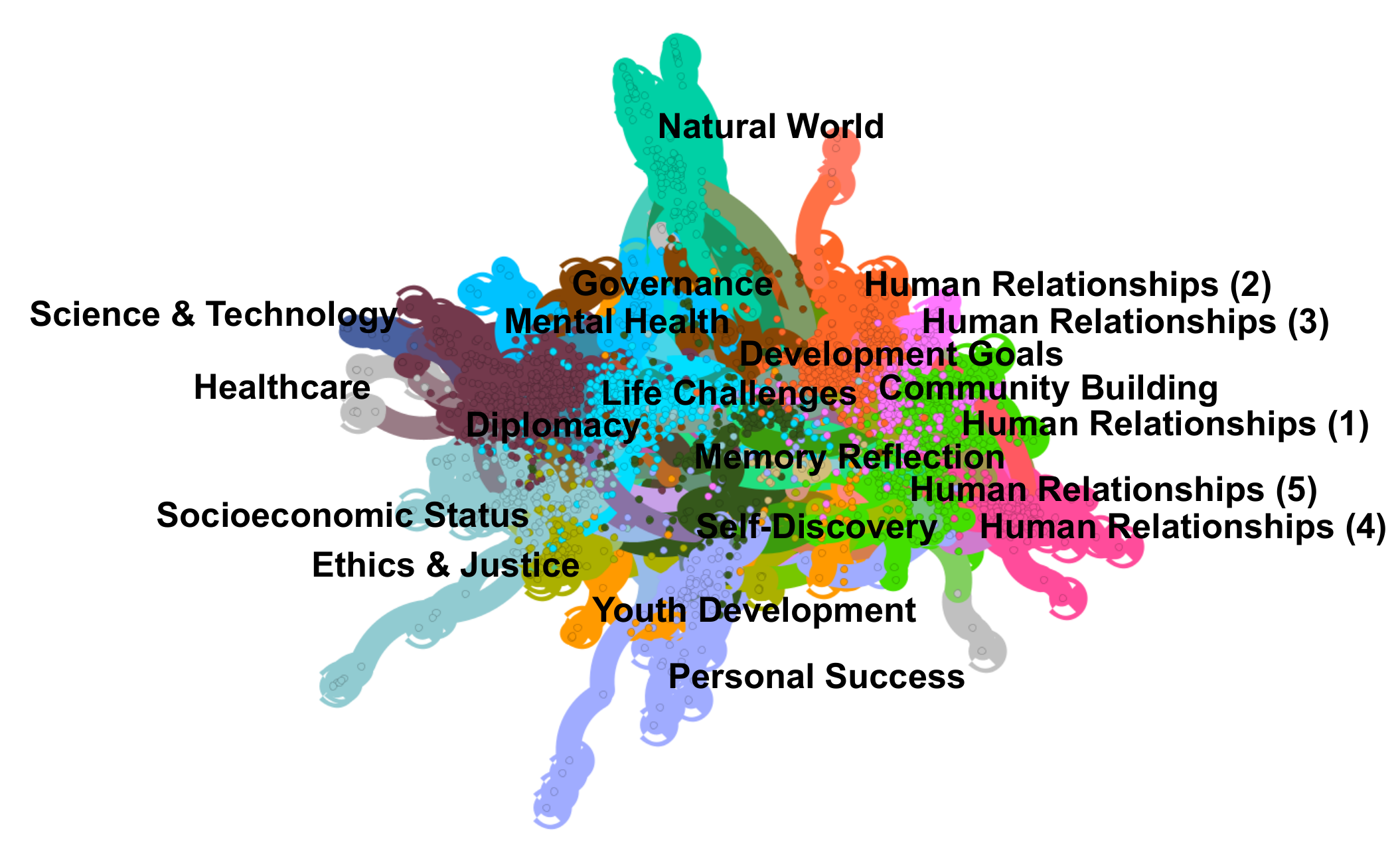}
  \caption{Variety of topics identified in our data.}
  \label{fig:topics_named}
\end{figure}

This section contains details on the topic analysis. Figure~\ref{fig:topics_named} presents the co-location of major topics discovered with a method of large network unfolding \citep{blondel2008fast} implemented in Gephi\footnote{\href{https://gephi.org/}{https://gephi.org/}}. We determined topics using this graph clustering technique to allow for more transparency and clear visualization of connections between topics, which is often questionable and may lack interpretation with more traditional topic models. The following prompt to GPT-3.5 was used to name them: ``\textit{Describe shortly the topic of the following group of utterances (in 10 words maximum): <list of 20 utterances from the topic>}''. Successively, another prompt was used to generalize topics within broader categories: ``\textit{I am providing you with a list of topics about the emotion of joy. Assign a broad category not related to emotions for each topic (2 words maximum). Topics can be from the same category. How different are the categories covered? Topics: <list of topics>}''. Table~\ref{tab:topics} shows the topics and their rather diverse categories. Only five topics belong to the same category of ``Human Relationships'' while the remaining are distinct.

\begin{table}[!t]
  \centering
  \begin{adjustbox}{width=\columnwidth}
  \begin{tabular}{l|c}
  \hline
  \textbf{Topic} & \textbf{Broader Category}\\
  \hline
    \makecell[{{p{0.64\textwidth}}}]{	Relief, freedom, success, and triumph after overcoming adversity or conflict.	}	&	Life Challenges	\\
\makecell[{{p{0.64\textwidth}}}]{	Love, reunion, happiness, and emotional connection with a significant person.	}	&	Human Relationships	\\
\makecell[{{p{0.64\textwidth}}}]{	Reunions, joy, and emotional connections with loved ones or friends.	}	&	Human Relationships	\\
\makecell[{{p{0.64\textwidth}}}]{	Love, reunion, joy, and deep connection in romantic relationships.	}	&	Human Relationships	\\
\makecell[{{p{0.64\textwidth}}}]{	Success, excitement, creativity, and joy in performances or achievements.	}	&	Personal Success	\\
\makecell[{{p{0.64\textwidth}}}]{	Triumph, achievement, power, wealth, and social or personal success.	}	&	Socioeconomic Status	\\
\makecell[{{p{0.64\textwidth}}}]{	Celebration of love, relationships, and happiness in couples' unions.	}	&	Human Relationships	\\
\makecell[{{p{0.64\textwidth}}}]{	Appreciation of natural beauty, peace, and uplifting moments in nature.	}	&	Natural World	\\
\makecell[{{p{0.64\textwidth}}}]{	Justice, accountability, and triumph over wrongdoing or injustice.	}	&	Ethics \& Justice	\\
\makecell[{{p{0.64\textwidth}}}]{	Joy, freedom, and feeling fully alive in transformative experiences.	}	&	Self-Discovery	\\
\makecell[{{p{0.64\textwidth}}}]{	Marriage, family, pregnancy, and joyful milestones in personal relationships.	}	&	Human Relationships	\\
\makecell[{{p{0.64\textwidth}}}]{	Achieving peace, quiet, and relief from stress or conflict.	}	&	Mental Health	\\
\makecell[{{p{0.64\textwidth}}}]{	Achieving important milestones like securing loans, building, or starting projects.	}	&	Development Goals	\\
\makecell[{{p{0.64\textwidth}}}]{	Nostalgia for simpler, happier times before loss or change occurred.	}	&	Memory Reflection	\\
\makecell[{{p{0.64\textwidth}}}]{	Achieving long-awaited success, freedom, or reunion, often involving royal matters.	}	&	Governance	\\
\makecell[{{p{0.64\textwidth}}}]{	Peaceful resolutions, strategic victories, and alliances in interstellar conflicts.	}	&	Diplomacy	\\
\makecell[{{p{0.64\textwidth}}}]{	Young people showing dedication, pride in others, and new discoveries.	}	&	Youth Development	\\
\makecell[{{p{0.64\textwidth}}}]{	Scientific breakthroughs leading to new weapons and peaceful advancements.	}	&	Science \& Technology	\\
\makecell[{{p{0.64\textwidth}}}]{	Collaboration, unity, and community support in achieving common goals.	}	&	Community Building	\\
\makecell[{{p{0.64\textwidth}}}]{	Successful surgeries, recoveries, and achievements in medical procedures and outcomes.	}	&	Healthcare	\\
    \hline
  \end{tabular}
  \end{adjustbox}
  \caption{Topics of the generated utterances for the emotion of \textit{joy} grounded in story plots.}
  \label{tab:topics}
\end{table}

\section{Human Evaluation Details}
\label{sec:appendix_human_evaluation}

\subsection{Detailed description of the task}

\begin{figure*}[ht]
  \includegraphics[width=\textwidth]{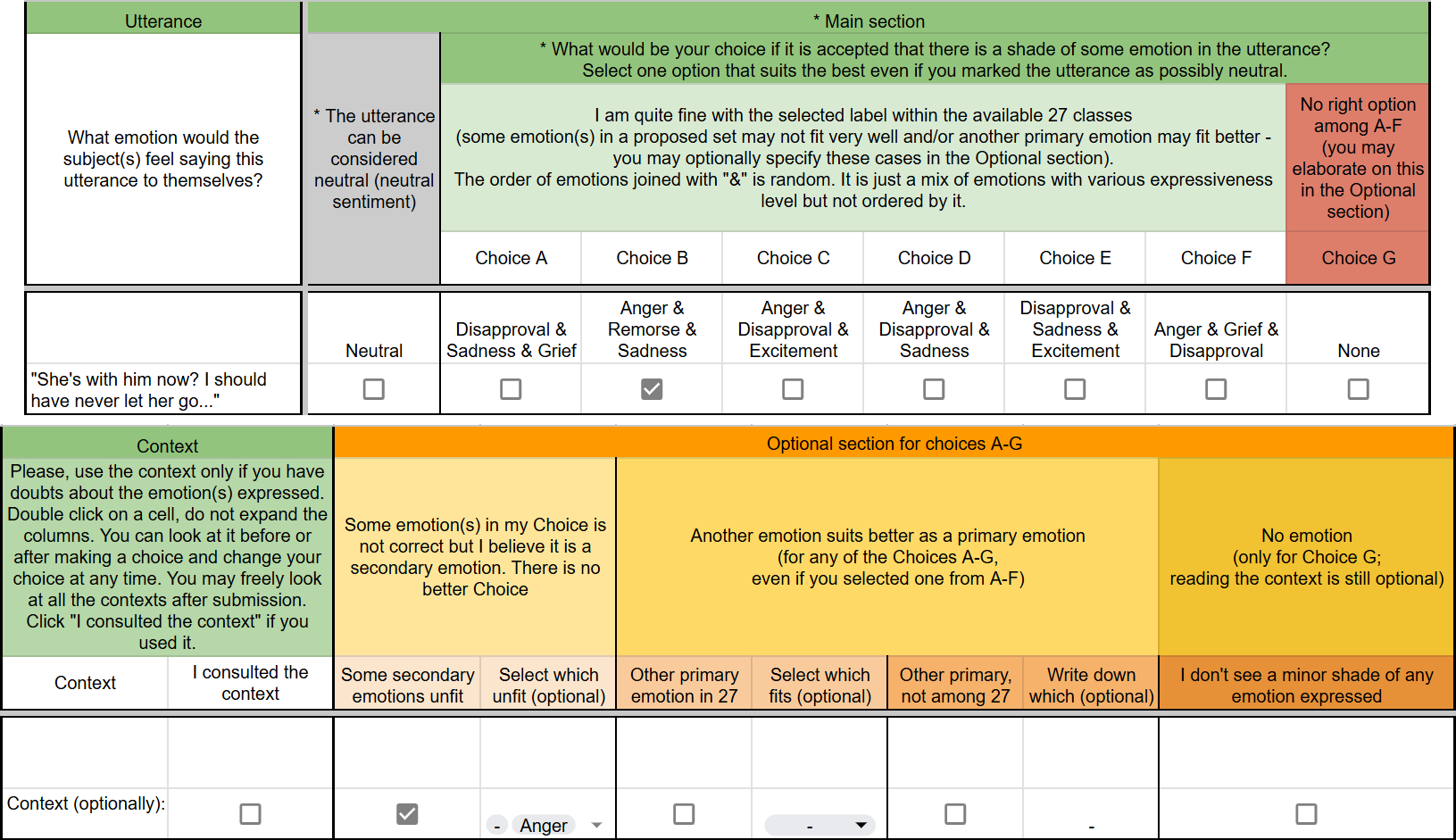}
  \caption{Human evaluation form.}
  \label{fig:evaluation_form}
\end{figure*}

The main part of the evaluation was designed in the form of a multiple-choice task due to the shown success in human annotator agreement in a similar emotion annotation task conducted in this form in \citet{sabour-etal-2024-emobench} that contrasts with a very low agreement when emotions are directly selected from the overall defined taxonomy as in \citet{demszky2020goemotions}. For each example, our setup offers several plausible choices, a few less plausible choices, and a \textit{none} option (7 in total). The context can be optionally viewed (it is hidden by default). The main part also asks whether the utterance could be considered neutral even if a \textit{non-none} option was chosen. We allowed for this flexibility, assuming that people may perceive whether an utterance is neutral in very different ways. Additionally, the optional part asks one to suggest missing or unfit emotions. The annotators were provided with definitions of emotions and informed that the utterances and contexts were generated automatically. To eliminate any potential biases, the distribution of emotions in the dataset and other details of the generation process remained unknown to the annotators. The evaluation form, along with an annotated example, is provided in Figure~\ref{fig:evaluation_form}. The task can be summarized with the following example:

\begin{quote}
Utterance: ``Look at them, they're so clumsy! I bet they'll make quite an impression tonight.''

Question 1: Can the utterance be considered neutral (neutral sentiment)?

Question 2: What emotion would the subject(s) feel saying this utterance to themselves?

Choices:
A)~Curiosity
B)~Surprise
C)~Amusement
D)~Approval
E)~Realization
F)~Annoyance
G)~None -- another emotion is better suited (indicate which, optionally).

[Label in the dataset to be validated (not visible to annotators): ``C''.]
\end{quote}

Instead of manually crafting options for every example as done in \citet{sabour-etal-2024-emobench}, we designed the following automatic procedure. For each label to be validated (note that it can be a set of labels like ``Pride \& Love \& Admiration''\footnote{In case there is more than one emotion assigned within the pipeline, we select at most three emotions with the largest expressiveness so as not to over-complicate the validation task.}), we randomly chose up to three additional emotions (up to five labels in total) from the emotional group(s) covered by the label to ensure more plausible options (the groups are shown in the header of Table~\ref{markers-table}). Successively, we also choose a few random emotions until we get a six-emotion set. We sample five various options from the compiled emotion set and shuffle them together with the dataset option. Emotions inside each of the six resulting options (including the one being validated) are also shuffled; their number in each option within an example is equal. The \textit{neutral} label does not appear among these six options. In case the dataset label is \textit{neutral}, it would correspond to the \textit{none} option while its spot within the first six options would be taken by another sampled label. This procedure successfully passed the approbation: according to the feedback from the annotators, there were only very few cases where the choice would be simplified due to a small number of plausible options.

We sampled 200 examples from the training set for the \textit{eval} set so that the number of \textit{neutral} labels is upsampled (to 10\% among all soft labels that made 11\% purely \textit{neutral} examples and 18\% with a \textit{neutral} label among others) to have a solid number that would allow us to reach proper conclusions on the neutrality aspect. We did not shuffle emotions in the validated option for the first 20\% of examples to check whether the generated expressiveness levels are valid and provide a helpful ranking (the annotators were only aware of the random order; see Figure~\ref{fig:evaluation_form} for exact instructions). Compound labels made 73\% of examples.

\subsection{Result Details}
\paragraph{Emotion label accuracy.} We calculated the accuracy of the dataset labels based on a set of examples where no annotator selected a neutral label as a possible class (i.e., on 64\% of the \textit{eval} set). The inter-annotator agreement using Cohen’s Kappa
\citep{kohen1960coefficient} is $\kappa = 0.365$, which points to the high task subjectivity, even though it is higher than $\kappa = 0.293$ in \citet{demszky2020goemotions}. The accuracy of our labels is 0.86 and 0.7 on examples where all three or at least two votes coincide, respectively.

All annotators noticed that in many examples, all the options had at least one emotion that did not fit the utterance. They still selected the best-suited choice, and in some cases, specified an irrelevant emotion. Manual analysis of these cases showed that inappropriate emotions appeared due to the mapping step introduced to replace out-of-taxonomy emotion labels generated by Mistral, as discussed in Section~\ref{sec:dataset_generation}. Additional experiments are required to confirm that this ``noise'' in the labels played a negative rather than a positive role in model training. We release original labels generated by Mistral along with the mapping.

\paragraph{Neutrality.} The number of examples annotated as purely \textit{neutral} by each annotator is: a) 6, b) 47 (covers 6/6 of \textit{`a)'}, i.e. 100\%), and c) 61 (covers 100\% of \textit{`a)'} and 89\% of \textit{`b)'}). The large difference in the number of selected examples confirms our assumption about the high subjectivity in the perception of neutrality. Both `\textit{b)}' and `\textit{c)}' selections covered the same 19 out of 21 (90\%) of the pure neutral dataset labels; the 6 of `\textit{a)}' also fully fall within these 19 choices. The high recall suggests that the pipeline can generate emotionless utterances. In verbal feedback, the annotators noticed that these examples lacked semantic variety, which coincides with our finding in Section~\ref{sec:data_analysis}.
\paragraph{Expressiveness.}
Some annotators mentioned that choosing emotions was somehow easier in the first part of the set. We calculated the accuracy scores per annotator for 20\% of the set where emotions in the validated option were ranked according to their expressiveness level, and separately for the remaining 80\% with the random order. The scores are shown in Table~\ref{tab:eval_expressiveness}.

\begin{table}[ht]
    \centering
     \begin{adjustbox}{width=\columnwidth}
    \begin{tabular}{l|c|c|c}
    \hline
         \textbf{Annotator} & \textbf{Ranked emotions} & \textbf{Shuffled emotions} & \textbf{Total}
         \\
         \hline
         A & 80.77 & 49.02 & 55.37\\
         B & 46.15 & 52.94 & 51.58\\
         C & 61.54 & 51.96 & 53.88\\ \hline
         \textbf{Average} & 62.82 & 51.31 & 53.61\\
    \hline
    \end{tabular}
    \end{adjustbox}
    \caption{Evaluation on two \textit{eval} subsets -- with and without ranking emotions by expressiveness inside the validated label. \textit{Total} is the weighted average.}
    \label{tab:eval_expressiveness}
\end{table}

The results suggest that the order helped two annotators perceive the emotion lists and made their choices simpler. This implicitly validates the correctness of the ranking and, thus, the values of expressiveness levels generated within the pipeline.

\paragraph{Context relevance.}
Checking the context was optional for annotators (it was hidden, and only double-clicking on the word ``context'' would open it). After opening it, the annotators had to tick the ``I consulted the context'' checkbox. This procedure was done to examine how important context was considered in clarifying emotions. According to the annotators' feedback, most contexts are helpful, but some are too long and/or too convoluted to make a judgment, or they do not help disambiguate emotions when the utterances are too short. No annotator complained about grammar or errors in the logical flow. This confirms that the ``Be as concise as possible.'' addition to the prompt was useful; however, some additional measures are required for those few cases when the model still summarizes the text with excessive information.

We also calculated that, on average, 77\% of cases where the context was consulted were when the \textit{rewritten} utterance was provided (precisely, 67\%, 71\% and 92\% per annotator). This justifies the success of our effort to make utterances less emotional, giving more importance to the context. However, as noticed above, in some cases, utterances become so ambiguous that even context does not help. More analysis is needed to verify that such examples still contribute positively to the training of the models and make them more robust to real-world inputs.

\begin{table*}
  \centering
  \begin{adjustbox}{width=\textwidth}
  \begin{tabular}{lll}
    \hline
    \textbf{Action}           & \textbf{Prompt} & \textbf{Outcome} \\
    \hline
    \makecell[{{p{0.1\textwidth}}}]{Extract \\actors}       & \makecell[{{p{0.51\textwidth}}}]{Plot: <text of the plot> \\ \\ Who are the characters in the plot? \\ Try to list all of them, one per line.} &   \makecell[{{p{0.66\textwidth}}}]{1. Rick Deckard (ex-police officer and Blade Runner) \\
2. Officer Gaff \\
3. Supervisor Bryant \\
4. Leon (Nexus-6 replicant) \\
5. Roy Batty (Nexus-6 replicant) \\
6. Zhora (Pris' companion and replicant) \\
7. Pris (Nexus-6 replicant) \\
8. Dr. Eldon Tyrell (CEO of Tyrell Corporation) \\
9. Rachael (experimental replicant) \\
...\\
16. Crowd members (background characters)}                        \\
    \hline
    \makecell[{{p{0.1\textwidth}}}]{Generate emotional utterances}     & \makecell[{{p{0.51\textwidth}}}]{Plot: <text of the plot> \\ Available list of emotions: \\ 
    <emotions and their definitions> \\ Actor: <extracted actor>
\\ \\
Generate 8 possible utterances of this actor thinking aloud that express  8 various non-neutral emotions according to the context in the plot. Additionally, generate 2 neutral utterances of this actor thinking aloud afterwards.} &  \makecell[{{p{0.66\textwidth}}}]{1. (Anger) "How could they send me after Rachael? She's not a replicant, she's human! I won't let Bryant or anyone else hurt her." \\ 
2. (Curiosity) "What's going on at the Tyrell Corporation? Why are these replicants here and what do they want from Tyrell?" \\
3. (Fear) "I've got Leon cornered, but he's so fast and strong. What if I can't retire him in time?" \\
...\\
8. (Optimism) "Maybe there's a way to save these replicants, to give them the chance to live beyond their four-year lifespan. I have to find a solution before it's too late." \\
Neutral: \\
1. "I need to focus, to find the replicants and retire them before they cause any more damage." \\
2. "I need to gather more information, to understand what's really going on and how best to approach this situation."          }               \\
    \hline
    \makecell[{{p{0.1\textwidth}}}]{Generate \\soft labels}       & \makecell[{{p{0.51\textwidth}}}]{The only possible list of emotions with their definitions: <classes (incl. neutral) and their definitions>\\ \\
    Select from the list above the top 5 emotions the utterance expresses. List them with an expressiveness level from 0 to 1 with a step of 0.1.\\ \\ Utterance: <utterance> \\ \\
    Start your response with: "1. <primary emotion>" and then add the following emotions with their expressiveness levels:\\ \\
1. <primary emotion>}        &   \makecell[{{p{0.66\textwidth}}}]{1. anger (1.0) - The speaker expresses strong feelings of displeasureand antagonism towards Bryant and others for sending  him after Rachael, who is perceived as innocent and human.\\
2. caring (1.0) - The speaker displays strong concern and kindness towards Rachael, expressing a desire to protect her from harm.\\
3. confusion (0.5) - The speaker seems puzzled or uncertain as to why Rachael is being targeted as a replicant.\\
4. desire (0.8) - The speaker expresses a strong desire to prevent harm from coming to Rachael.\\
5. neutral (0.1) - The speaker's tone and language do not indicate any particular expressiveness for the remaining emotion classes.}                        \\
    \hline
    \makecell[{{p{0.1\textwidth}}}]{Generate context}       & \makecell[{{p{0.51\textwidth}}}]{Plot: <text of the plot> \\ Actor: <extracted actor> \\ Actor's utterance: <utterance> \\ Expressed emotions: <emotions> \\ \\
    Explain why the actor was thinking aloud this way starting from as close to the beginning of the story as needed to provide a complete picture but only until the moment of the utterance. Avoid talking about the given emotional state of the actor throughout the explanation. Be as concise as possible. }        &   \makecell[{{p{0.66\textwidth}}}]{Rick Deckard was initially reluctant to hunt down the replicants, including Rachael, after being informed by his supervisor Bryant that they had come to Earth illegally. However, after being threatened ambiguously by Bryant, Deckard agreed to retire them. During his investigation, he discovered that Rachael was an experimental replicant who believed herself to be human, with implanted false memories. This revelation led Deckard to question the validity of the Voight-Kampff test and the distinction between replicants and humans. When he encountered Rachael in person, he was moved by her emotional response and began to doubt her status as a replicant. The realization that she was in danger from the other replicants and Bryant fueled his determination to protect her, leading him to declare that she was human and not a replicant, despite the evidence to the contrary.}                        \\
    \hline
    \makecell[{{p{0.1\textwidth}}}]{Clean context from emotions}      & \makecell[{{p{0.51\textwidth}}}]{Character: <extracted actor> \\ Context: <generated context> \\
   \\ Remove clauses or even entire sentences \\from the context that explicitly discuss the \\emotions of <emotions> in the character.}        &   \makecell[{{p{0.66\textwidth}}}]{Rick Deckard was initially hesitant to retire the illegal replicants, including Rachael, as he was informed of their presence on Earth by his supervisor Bryant. \\...\\ 
    Despite the evidence suggesting otherwise, Deckard declared Rachael to be human.}                        \\
    \hline
    \makecell[{{p{0.1\textwidth}}}]{Rewrite utterance}      & \makecell[{{p{0.51\textwidth}}}]{Summary: <cleaned context>\\Character: <extracted actor>\\Character's utterance: <utterance>\\Expressed emotions: <emotions>\\ \\
    Rewrite the utterance so that the emotions are ambiguous without the summary. Be as concise as possible.}        &   \makecell[{{p{0.66\textwidth}}}]{How could they ask me to target Rachael? She's not what I expected. I won't let anyone harm her.}                        \\
    \hline
  \end{tabular}
  \end{adjustbox}
  \caption{\label{prompts-table}
    Prompts used in our synthetic data generation pipeline and corresponding outcomes.
  }
\end{table*}

\end{document}